\documentclass{article}

    \usepackage[preprint]{neurips_2021}

\usepackage[utf8]{inputenc} % allow utf-8 input
\usepackage[T1]{fontenc}    % use 8-bit T1 fonts
\usepackage{hyperref}       % hyperlinks
\usepackage{url}            % simple URL typesetting
\usepackage{booktabs}       % professional-quality tables
\usepackage{amsfonts}       % blackboard math symbols
\usepackage{nicefrac}       % compact symbols for 1/2, etc.
\usepackage{microtype}      % microtypography

\usepackage{natbib}
\usepackage{amsmath,amsfonts}
\usepackage[pdftex]{graphicx}
\usepackage{subfigure}
\usepackage{caption, floatrow} % 
\usepackage{multirow}

\usepackage[table,xcdraw]{xcolor}
\definecolor{LightCyan}{rgb}{0.88,1,1}
\definecolor{hotpink}{RGB}{255,105,180}
\definecolor{deepskyblue}{RGB}{0,191,255}

\newcommand{\litecap}[1]{\textcolor{deepskyblue}{#1}}
\newcommand{\pinkcap}[1]{\textcolor{hotpink}{#1}}

\def\ie{{\em i.e.},\ }

\def\eg{{\em e.g.},\ }

\def\cf{{\em cf.},\ }

\def\fn#1#2{#1(\, #2\,)}
\def\cfn#1#2#3{\fn{#1}{#2\,|\,#3}}
\def\Pr#1{\fn{\textrm{Pr}}{#1}}
\def\CPr#1#2{\cfn{\textrm{Pr}}{#1}{#2}}

\def\KL#1#2{\textrm{KL}\big(\, #1\ ||\ #2 \, \big)}

\long\def\comment#1{}

\newcommand\independent{\protect\mathpalette{\protect\independenT}{\perp}}
\def\independenT#1#2{\mathrel{\rlap{$#1#2$}\mkern2mu{#1#2}}}
\def\Real{{\cal R}}
\title{Variational Auto-Encoder Architectures \\ that Excel at Causal Inference}

\author{%
  Negar Hassanpour \\
  Department of Computing Science\\
  University of Alberta, Amii\\
  Edmonton, Canada \\
  \texttt{hassanpo@ualberta.ca} \\
  \And
  Russell Greiner \\
  Department of Computing Science\\
  University of Alberta, Amii\\
  Edmonton, Canada \\
  \texttt{rgreiner@ualberta.ca} \\
}

\begin{document}

\maketitle

\begin{abstract}
    Estimating causal effects from observational data (at either an individual- or a population- level)
    is critical for making many types of decisions.
    One approach to address this task is to learn decomposed representations of the underlying factors of data;
    this becomes significantly more challenging when there are confounding factors (which influence both the cause and the effect).
    In this paper, we take a generative approach that 
    builds on the recent advances in Variational Auto-Encoders
    to simultaneously learn those underlying factors
    as well as the causal effects.
    We propose a progressive sequence of models, 
    where each improves over the previous one,
    culminating in the Hybrid model. 
    Our empirical results demonstrate that the performance of all three proposed models are superior to 
    both state-of-the-art discriminative as well as other generative approaches in the literature. 
\end{abstract}

\section{Introduction}
As one of the main tasks in studying causality~\citep{peters2017elements,guo2018survey},
the goal of Causal Inference is to 
determine % figure out
\emph{how much} the value of a certain variable would change
(\ie the \textbf{effect}) had another specified variable 
(\ie the \underline{cause}) 
changed its value.
A prominent example is the counterfactual question~\citep{rubin1974estimating,pearl2009causality}
``Would this patient have \textbf{lived longer} 
(and by \emph{how much}), 
had she received an alternative \underline{treatment}?''.
Such questions are often asked in the context of precision medicine,
% --- \ie the customization of health-care tailored to each individual patient --- 
hoping to identify which medical procedure~$t \in \mathcal{T}$ will benefit a certain patient~$x$ the most,
in terms of the treatment outcome~$y \in \Real$ (\eg survival time).

The first challenge with causal inference is the \textbf{unobservability} \citep{holland1986statistics} of the counterfactual outcomes
(\ie outcomes pertaining to the treatments that were {\em not}\ administered).
In other words, the true causal effect is never observed %(\ie missing in any training data)
and cannot be used to train predictive models, 
nor can it be used to evaluate a proposed model.
% This makes estimating causal effects a more difficult problem than that of generalization in the supervised learning paradigm.
The second common challenge is that 
the training data is often an observational study that exhibits \textbf{selection bias}~\citep{imbens_rubin_2015}
--- \ie the treatment assignment can depend on the subjects' attributes.
% (as opposed to a Randomized Controlled Trial (RCT) setting).
The general problem setup of causal inference and its challenges are described in detail in Appendix~\ref{app:ci}.

Like any other machine learning task, we can employ either of the two general approaches to address the problem of causal inference:
(i)~discriminative modeling, 
or (ii)~generative modeling, 
which differ in how the input features $x$ and their target values $y$ are modeled~\citep{ng2002discriminative}:

\textbf{Discriminative methods} focus solely on modeling the conditional distribution $\CPr{y}{x}$ with the goal of direct prediction of the target $y$ for each instance $x$. 
For prediction tasks, discriminative approaches are often more accurate since they use the model parameters more efficiently than generative approaches.
Most of the current causal inference approaches are discriminative, 
including the matching-based methods such as
% Deep Match \citep{kallus2020deepmatch} and
Counterfactual Propagation \citep{harada2020counterfactual},
as well as 
the regression-based methods such as
Balancing Neural Network (BNN) \citep{johansson2016learning},
CounterFactual Regression Network (CFR-Net) \citep{shalit17a}
and its extensions 
(\cf \citep{yao2018representation,hassanpour2019counterfactual}), and
% Similarity preserved Individual Treatment Effect (SITE) \citep{yao2018representation}, and
Dragon-Net \citep{shi2019adapting}.

\begin{figure}[t]
% %%\vskip -.75cm
\centering
    \begin{minipage}[b]{0.45\linewidth}
        \centering
        \includegraphics[scale=.45]{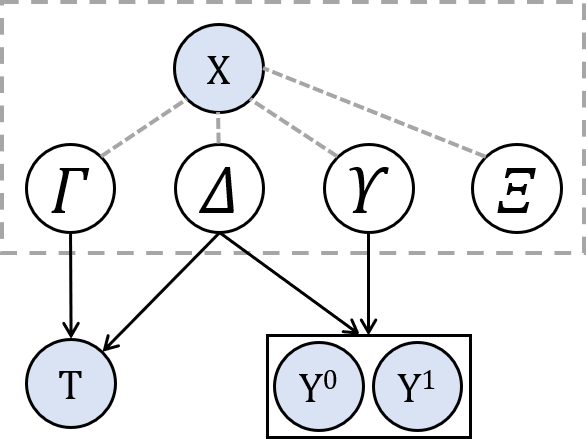}
        \caption{The three underlying factors of an observational data (namely $\Gamma$, $\Delta$, and $\Upsilon$) \citep{hassanpour2020learning}.
        $\Xi$ represents noise.}
        \label{fig:srcs}
    \end{minipage}
\end{figure}

\textbf{Generative methods}, on the other hand, 
describe the relationship between $x$ and $y$ by their joint probability distribution $\Pr{x, y}$.
This, in turn, allows the generative model to answer arbitrary queries,
including coping with
missing features using the marginal distribution $\Pr{x}$
or
    [similar to discriminative models]
    predicting the unknown target values $y$ via $\CPr{y}{x}$.  
A promising direction forward for causal inference is developing \emph{generative} models,
using either
    Generative Adverserial Network (GAN) \citep{goodfellow2014generative}
or 
    Variational Auto-Encoder (VAE) \citep{kingma2014auto,rezende2014stochastic}.
This has led to three generative approaches for causal inference: 
GANs for inference of Individualised Treatment Effects (GANITE) \citep{yoon2018ganite}, 
Causal Effect VAE (CEVAE) \citep{louizos2017causal}, and
Treatment Effect by Disentangled VAE (TEDVAE) \citep{zhang2021treatment}.
% However, neither GANITE nor CEVAE achieve competitive performance in terms of treatment effect estimation compared to the discriminative approaches.
% Although TEDVAE achieves competitive performance, it fails to fully disentangle the underlying factors of observational data (see Figure~\ref{fig:srcs}).
However, these generative methods either do not achieve competitive performance compared to the discriminative approaches or 
come short of fully disentangling the underlying factors of observational data (see Figure~\ref{fig:srcs}).

Although discriminative models have excellent predictive performance, 
they often suffer from two drawbacks:
(i)~overfitting, 
and
(ii)~making highly-confident predictions, even for 
instances that are ``far'' 
from the observed training data.
Generative models based on Bayesian inference, on the other hand, can handle both of these drawbacks:
issue~(i) can be 
minimized % avoided
by taking an average over the posterior distribution of model parameters;
and
issue~(ii) can be 
addressed % overcome
by explicitly 
providing % accounting for
model uncertainty via the posterior~\citep{gordon2020combining}.
Although  the exact inference is often intractable, 
efficient approximations to the parameter posterior distribution is possible through variational methods.
In this work, we use the Variational Auto-Encoder (VAE) framework \citep{kingma2014auto,rezende2014stochastic} to tackle this.

\textbf{Contributions:} 
We propose three interrelated Bayesian models (namely Series, Parallel, and Hybrid) 
% ---  built on top of each other ---
that employ the VAE framework to address the task of causal inference for binary treatments.
We demonstrate that all three of these models significantly outperform the state-of-the-art 
in terms of estimating treatment effects
on two publicly available benchmarks, 
as well as a fully synthetic dataset that allows for detailed performance analyses.
We also show that our proposed Hybrid model is the best at decomposing the underlying factors of any observational dataset.
% This is a valuable property, 
% as that means it can accurately estimate all treatment outcomes.

% \add[RG]
% \comment
{
The rest of this document is organized as follows:
Section~\ref{sec:back} provides the background and related work; 
Section~\ref{sec:method} elaborates on the proposed method;
Section~\ref{sec:exp} reports and discusses the experimental results; and
Section~\ref{sec:conc} concludes the paper with future directions and a summary of contributions.
}

\section{Related Works} %2
\label{sec:back}
% \add[RG]
{For notation, we will use $x$ to describe an instance,
and let $t \in \{0,1\}$ refer to the treatment administered, yielding the outcome with value $y \in \Real$;
see Appendix~\hbox{\ref{app:ci}} for details.}

% \noindent 
\textbf{CFR-Net} $\quad$
\citet{shalit17a} considered the binary treatment task and 
attempted to learn a representation space $\Phi$ that reduces selection bias by  
making $\CPr{\Phi(x)}{t\!=\!0}$ and $\CPr{\Phi(x)}{t\!=\!1}$ as close to each other as possible,
% (see Figure~\ref{fig:rep}),
provided that $\Phi(\,x\,)$ retains enough information that 
the learned regressors 
$\{\, h^t\big( \Phi(\cdot) \big)\,: %, \; \forall 
\ t \! \in\! \{0, 1\}\, \}$ 
can generalize well on the observed outcomes.
Their objective function includes 
$
    \ L\big[\, y_i,\, h^{t_i}\big( \Phi( x_i )\big) \big]
$,
which is the loss of predicting the observed outcome for 
instance % sample
$i$ (described as $x_i$), 
weighted by 
$
    \omega_i = \frac{t_i}{2u} + \frac{1-t_i}{2(1-u)} 
$,
where 
$
    u = \Pr{t\!=\!1}
$.
This is effectively setting
$
    \omega_i
    = \frac{1}{2\,\Pr{t_i}}
$
where $\Pr{t_i}$ is the probability of 
selecting treatment $t_i$ over the entire population.

\textbf{DR-CFR} $\quad$
\citet{hassanpour2020learning} argued against the standard implicit assumption that \emph{all} of the covariates $X$ are confounders
(\ie contributing to both treatment assignment and outcome determination).
Instead, they proposed the graphical model shown in Figure~\ref{fig:srcs} 
(with the underlying factors $\Gamma$, $\Delta$, and $\Upsilon$)
and designed a 
{\em discriminative}\ causal inference approach accordingly.
% --- built on top of the CFR-Net.
Specifically, their ``Disentangled Representations for CFR'' ({DR-CFR}) model 
includes three representation networks,
each trained with constraints to ensure that each component corresponds to its respective underlying factor.
While the idea behind {DR-CFR} provides an interesting intuition,
it is known that only generative models 
(and not discriminative ones) 
can truly identify the underlying data generating mechanism.
This paper is a step in this direction.

\textbf{Dragon-Net} $\quad$
\citet{shi2019adapting}'s main objective was to estimate the Average Treatment Effect (ATE),
which they explain requires a two stage procedure:
(i)~fit models that predict the outcomes for each treatment;
and
(ii)~find a downstream estimator of the effect.
Their method is based on a classic result from strong ignorability 
(\ie Theorem~3 in \citep{rosenbaum1983central}) 
that states:
\begin{align*}
    &(y^1, y^0) \independent t \,|\, x      &\&&& \quad \CPr{t=1}{x}      &\in (0,1) &\implies \\
    &(y^1, y^0) \independent t \,|\, b(x)   &\&&& \quad \CPr{t=1}{b(x)}   &\in (0,1)
\end{align*}
where $b(x)$ is a balancing score%
\footnote{
    That is, $X \independent T \,|\, b(X)$\quad \citep{rosenbaum1983central}.
}
(here, propensity)
and
argued that only the parts of $X$ relevant for predicting $T$
are required for the estimation of the causal effect.%
\footnote{
    The authors acknowledge that this would hurt the predictive performance for individual outcomes.
    As a result, this yields inaccurate estimation of Individual Treatment Effects (ITEs).
}
This theorem only provides a way to \emph{match} treated and control instances though 
--- \ie it helps finding potential counterfactuals from the alternative group,
% \add[RG]
{which they use}
to calculate ATE.
\citet{shi2019adapting}, however, used this theorem to derive minimal representations on which to \emph{regress} to estimate the outcomes.
% Clearly, if true, this requires a proof.

\textbf{GANITE} $\quad$
\label{sec:ganite}
\citet{yoon2018ganite} proposed the counterfactual GAN,
whose generator $G$, 
given $\{ x, t, y^t \}$, 
estimates the counterfactual outcomes ($\hat{y}^{\neg t}$); 
and whose discriminator $D$ tries to identify 
which of $\{[x, 0, y^0],\ [x,1, y^1]\}$ is the factual outcome.
It is, however, 
unclear why this requires that $G$ must produce samples that are indistinguishable from the factual outcomes, 
especially as $D$ can just learn the 
\emph{treatment selection mechanism} 
% \add[RG]
{(\hbox{\ie} the mapping from $X$ to $T$)}
instead of distinguishing the factual outcomes from counterfactuals. 
Although this work is among the few generative approaches for causal inference, % in the literature,
our empirical results (in Section~\ref{sec:exp}) show that it does not effectively estimate counterfactual outcomes. 

\begin{figure}[t]
\quad \quad
    \begin{minipage}[b]{0.45\linewidth}
        \centering
        \includegraphics[scale=.45]{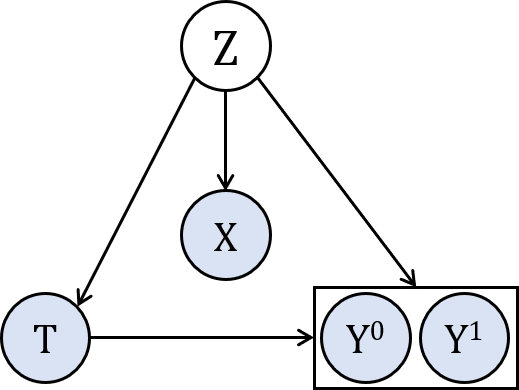}
        \caption{Graphical model of the CEVAE method \citep{louizos2017causal}}
        \label{fig:cevae}
    \end{minipage}
%%\vskip -.25cm
% \note{1. Legend for BOTH models.\\
% 2. This is fig 2. Where is Fig 1?}
\end{figure}

\textbf{CEVAE} $\quad$
\label{sec:cevae}
\citet{louizos2017causal} used VAE to extract latent confounders from their observed proxies in $X$.
While this is a step in the right direction,
empirical results show that it does not always accurately estimate treatment effects
(see Section~\ref{sec:exp}).
The authors note that this may be because
CEVAE is not able to address the problem of selection bias.
Another reason for CEVAE's sub-optimal performance might be 
its assumed graphical model of the underlying data generating mechanism, depicted in Figure~\ref{fig:cevae}.
This model assumes that there is only one latent variable $Z$ (confounding $T$ and $Y$) that generates the entire observational data;
however, \citep{kuang2017treatment,hassanpour2020learning} have shown the advantages of involving more
factors (see Figure~\ref{fig:srcs}).

\textbf{TEDVAE} $\quad$
Similar to \hbox{DR-CFR} \citep{hassanpour2020learning}, 
\citet{zhang2021treatment} proposed TEDVAE in an attempt to learn disentangled factors but using a \emph{generative} model instead
(\ie a VAE with a three-headed encoder, one for each underlying factor).
While their method proposed an interesting intuition on how to achieve this task, 
according to the reported empirical results (see their Figure~4c), 
% \change[RG]{it seems}
{the authors found} that TEDVAE is not successful in identifying the risk factors $z_y$ (equivalent to our $\Upsilon$).
This might be 
because their % due to the fact that the 
model does not have a mechanism 
for % of 
distinguishing between the risk factors and confoundings $z_c$ (equivalent to our $\Delta$).
The evidence is in their Equation~(8), 
% \change[RG]{where $z_y$ can be degenerate and all information be embedded in $z_c$}
{which would allow $z_y$ to be degenerate and have all information embedded in $z_c$}.
Our work, however, proposes % In this work, we propose 
an architecture that can achieve this decomposition.
\\
\\
\textbf{M1 and M2 VAEs} $\quad$
The M1 model~\citep{kingma2014auto} is the conventional VAE, 
which learns a latent representation from the covariate matrix $X$ alone in an unsupervised manner.
% \note[RG]{?? does it use this for this counterfactual task? Or ... ??}
% \note[NH]{M1 generates the representation, which may be used in any downstream task such as counterfactual regression.}
\citet{kingma2014semi} extended this to the M2 model 
that, % which,
in addition to the covariates,
also allows the target information to guide the representation learning process in a semi-supervised manner.
Stacking the M1 and M2 models
% \change[RG]{yield the}
{produced their} best results:
first learn a representation $Z_1$ from the raw covariates,
then find a second representation $Z_2$,
% \note[RG]{of just the covariates X?? How is that better? What about T?}
% \note[NH]{z1 and z2 learn representations from X and Y. For our task where we also have a second target T (in addition to Y), another M1+M2 model accounts for T, that is latent factors z3 and z4.}
now learning from $Z_1$ (instead of the raw data) as well as the target information. % \add[NH]{and guided by the target information}.
{Appendix~\ref{app:m1m2}}
presents a more detailed overview of the M1 and M2 VAEs.
In our work, the target information includes the treatment bit $T$ as well as the observed outcome $Y$.%
\footnote{
Therefore, we require multiple stacked models here.
}
This additional information helps the model to learn more expressive representations, 
which % that
was not possible with the unsupervised M1 model.

\section{Method}
\label{sec:method}
Following \citep{kuang2017treatment,hassanpour2020learning} and without loss of generality, 
we assume that 
% the random variable 
$X$ follows an unknown joint probability distribution 
$\CPr{X}{\Gamma, \Delta, \Upsilon, \Xi}$, 
where $\Gamma$, $\Delta$, $\Upsilon$, and $\Xi$ are non-overlapping {independent} factors.
Moreover, we assume that the treatment $T$ follows $\CPr{T}{\Gamma, \Delta}$ 
(\ie $\Gamma$ and $\Delta$ are responsible for selection bias)
and 
the outcome $Y^{^{T}}$ follows $\textrm{Pr}_{_{T}}(\, Y^{^{T}} \,|\, \Delta, \Upsilon \,)$
--- see Figure~\ref{fig:srcs}.
Observe that the factor $\Gamma$~(respectively,~$\Upsilon$) 
partially determines only $T$~(respectively,~$Y$),
but not $Y$~(respectively,~$T$); and
$\Delta$ includes the confounding factors between $T$ and $Y$.

We emphasize that the belief net in Figure~\ref{fig:srcs} 
is built without loss of generality;
\ie it also covers the scenarios where any of the latent factors is degenerate.
Therefore, if we can design a method that has the capacity to capture all of these latent factors,
it would be successful in all scenarios --- even in the ones that have degenerate factors
(and in fact this is true; 
see the experimental setting and results of the Synthetic benchmark in Sections~\ref{sec:bench}~and~\ref{sec:tee}).

\begin{figure*}[t]
%%\vskip -.75cm
    \begin{center}
        \subfigure[The \textbf{Series} Model]{\label{fig:series}\includegraphics[scale=.45]{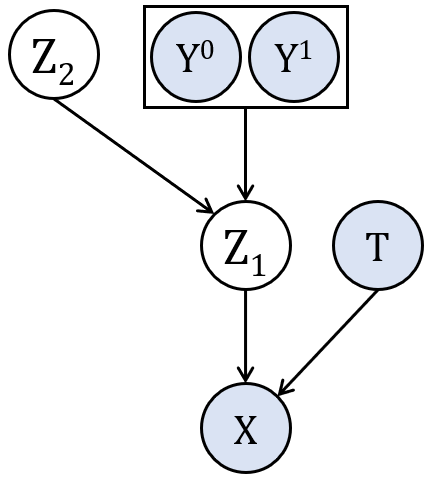}}
        \qquad \qquad 
        \subfigure[The \textbf{Parallel} Model]{\label{fig:parallel}\includegraphics[scale=.45]{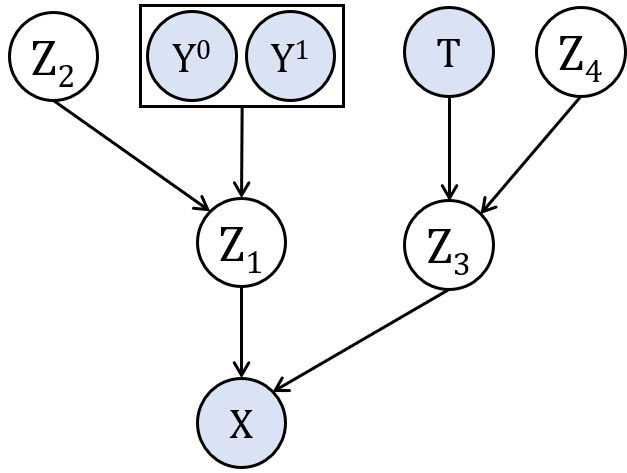}}
        \qquad \qquad 
        \subfigure[The \textbf{Hybrid} Model]{\label{fig:combo}\includegraphics[scale=.45]{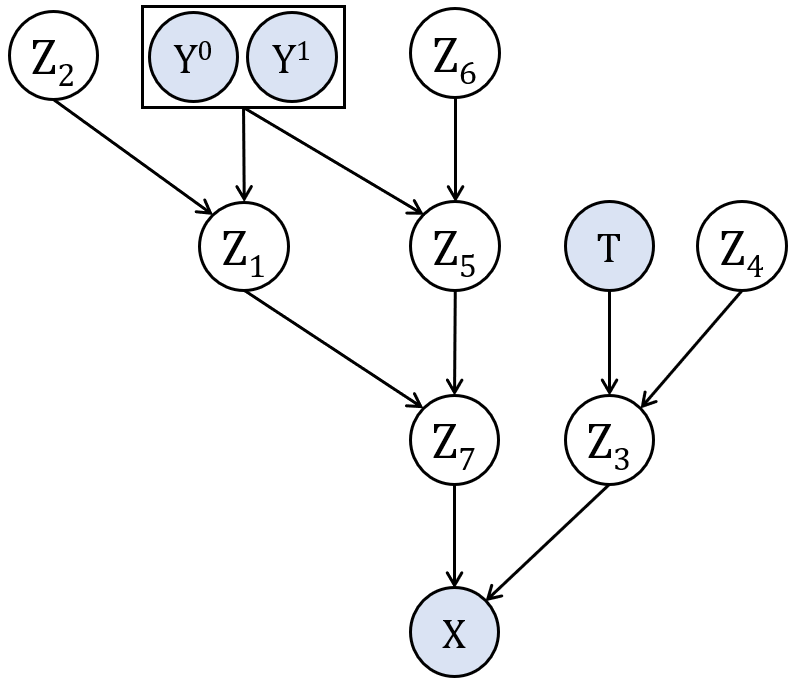}}
    \end{center}
%%\vskip -.5cm
    \caption{
        Belief nets of the proposed models.
    }
    \label{fig:models}
\end{figure*}

Our goal is to design a generative model architecture that 
encourages learning decomposed representations of these underlying latent factors 
(see Figure~\ref{fig:srcs}).
In other words, 
it should be able to decompose and separately learn the three underlying factors that are responsible for determining 
``$T$ only''~($\Gamma$), 
``$Y$ only''~($\Upsilon$), and
``both $T$ and $Y$''~($\Delta$).
To achieve this, we propose a progressive sequence of three models 
(namely Series, Parallel, and Hybrid; 
as illustrated in Figures~\ref{fig:series},~\ref{fig:parallel},~and~\ref{fig:combo} respectively), 
where each is an \emph{improvement} over the previous one.
Every model employs several stacked M1+M2 VAEs \citep{kingma2014semi}, 
that each includes
    a decoder (generative model)
and
    an encoder (variational posterior),
which are parametrized as deep neural networks.

\subsection{The Variational Auto-Encoder Component}
\subsubsection{The Series Model} % 3.A.1)
The belief net of the Series model is illustrated in Figure~\ref{fig:series}.
\citet{louizos2015variational} proposed a similar architecture to address fairness in machine learning,
but using a binary sensitive variable $S$ (\eg gender, race, etc.) rather than the treatment $T$.
Here, we employ this architecture for causal inference and explain why it should work. % in this setting.
% \add[R4]{
We hypothesize that this structure functions as a fractionating column%
\footnote{
In chemistry, a fractionating column is used for separating different liquid compounds in a mixture; 
see \url{https://en.wikipedia.org/wiki/Fractionating_column} for more details.
In our work, similarly, we can separate different factors from the pool of features using the proposed VAE architecture.
}:
% \note[RG]{What is a ref for "Distillation Tower"?} \note[NH]{I myself made up this metaphor (in English it's actually called fractionating column with which we can separate different liquid compounds). Here, similarly, we can separate different factors from the pool of features. It's a side effect of having a chemical engineer dad!! Not sure if the metaphor is too far-fetched...?}
the bottom M2 VAE attempts to decompose $\Gamma$~(guided by $T$) 
from $\Delta$ and $\Upsilon$~(captured by $Z_1$);
and the top M2 VAE attempts to learn $\Delta$ and $\Upsilon$~(guided by $Y$).
% }

The decoder and encoder components of the Series model 
--- $p(\cdot)$ and $q(\cdot)$ parametrized by $\theta_s$ and $\phi_s$ respectively ---
involve the following distributions:
\begin{table}[H]
\setlength{\tabcolsep}{2.5pt}
%\vskip -.2in 
    \begin{center}
        \begin{tabular}{ll|l|ll}
        \multicolumn{2}{c|}{\bf Priors}  &\multicolumn{1}{c|}{\bf Likelihood}  &\multicolumn{2}{c}{\bf Posteriors}    \\
        \hline
        $p_{\theta_s}(z_2)$ &$p_{\theta_s}(z_1 | y, z_2)$ &$p_{\theta_s}(x | z_1, t)$ &$q_{\phi_s}(z_1 | x, t)$ &$q_{\phi_s}(y | z_1)$ \\
        &&&$q_{\phi_s}(z_2 | y, z_1)$ \\
        \end{tabular}
    \end{center}
%\vskip -.3in 
\end{table}

\noindent Hereafter, we drop the $\theta$ and $\phi$ subscripts for brevity.

The goal is to maximize the conditional log-likelihood of the observed data 
(left-hand-side of the following inequality)
by maximizing the Evidence Lower BOund 
(ELBO; right-hand-side) % of the following inequality
--- \ie
\begin{align}
    \sum_{i=1}^{N} \log p(x_i | t_i, y_i) \quad \geq \quad
    \sum_{i=1}^{N} \mathbb{E}_{q(z_1 | x, t)} \big[ \log p(x_i | z_{1_i}, t_i) \big]  \label{eq:recl_s} \\
    -\KL{q(z_1 | x, t)}{p(z_1 | y, z_2)}
    -\KL{q(z_2 | y, z_1)}{p(z_2)} \label{eq:kld_s}
\end{align}
\comment{
\begin{eqnarray}
    \sum_{i=1}^{N} \log p(x_i | t_i, y_i) &\quad \geq &\quad
    \sum_{i=1}^{N} \mathbb{E}_{q(z_1 | x, t)} \big[ \log p(x_i | z_1_i, t_i) \big]  \label{eq:recl_s} \\
   &&-\quad\KL{q(z_1 | x, t)}{p(z_1 | y, z_2)} \label{eq:kld_s} \\
   &&  -\quad \KL{q(z_2 | y, z_1)}{p(z_2)} \nonumber
\end{eqnarray}
}
where KL denotes the Kullback-Leibler divergence,
$p(z_2)$ is the unit multivariate Gaussian
(\ie $\mathcal{N}(0, \mathbb{I})$), and
the other distributions are parameterized as deep neural networks.

\subsubsection{The Parallel Model}
% \add[R4]{
    The Series model is composed of two M2 stacked models.
    However, {\citet{kingma2014semi}} showed that an M1+M2 stacked architecture learns better representations than an M2 model alone 
    for a downstream prediction task.
    This motivated us to design a double M1+M2 Parallel model;
    where one arm is for the outcome to guide the representation learning via $Z_1$ 
    and another for the treatment to guide the representation learning via $Z_3$.
% }
Figure~\ref{fig:parallel} shows the belief net of this model.
We hypothesize that $Z_1$ would learn $\Delta$ and $\Upsilon$, and $Z_3$ would learn $\Gamma$ (and perhaps partially $\Delta$).

The decoder and encoder components of the Parallel model 
--- $p(\cdot)$ and $q(\cdot)$ parametrized by $\theta_p$ and $\phi_p$ respectively ---
involve the following distributions:
\begin{table}[H]
\setlength{\tabcolsep}{5pt}
%\vskip -.2in 
    \begin{center}
        \begin{tabular}{ll|l|ll}
        \multicolumn{2}{c|}{\bf Priors}  &\multicolumn{1}{c|}{\bf Likelihood}  &\multicolumn{2}{c}{\bf Posteriors}    \\
        \hline
        $p(z_2)$ & $p(z_1 | y, z_2)$ & $p(x | z_1, z_3)$ & $q(z_1 | x, t)$     & $q(y | z_1)$      \\
        $p(z_4)$ & $p(z_3 | t, z_4)$ &                   & $q(z_2 | y, z_1)$   &$q(t | z_3) $       \\
                 &                   &                   & $q(z_3 | x, y)$   &   \\
                 &                  &                   & $q(z_4 | t, z_3)$ &   \\
        \end{tabular}
    \end{center}
%\vskip -.3in 
\end{table}

Here, the conditional log-likelihood can be upper bounded by:
\begin{align}
    \sum_{i=1}^{N} \log p(x_i | t_i, y_i) \quad \geq \quad
    \sum_{i=1}^{N} \mathbb{E}_{q(z_1, z_3 | x, t, y)} \big[ \log p(x_i | z_{1_i}, z_{3_i}) \big] \label{eq:recl_p} \\
    -\KL{q(z_1 | x, t)}{p(z_1 | y, z_2)}
    -\KL{q(z_2 | y, z_1)}{p(z_2)}            \nonumber \\
    -\KL{q(z_3 | x, y)}{p(z_3 | t, z_4)}
    -\KL{q(z_4 | t, z_3)}{p(z_4)} \label{eq:kld_p} 
\end{align}

\subsubsection{The Hybrid Model}
% \add[R4]{
    The final model, Hybrid (see Figure~\ref{fig:combo}), attempts
    to combine the best capabilities of the previous two architectures.
    The backbone of the Hybrid model has a Series architecture, 
    that separates $\Gamma$ (factors related to the treatment $T$; captured by the right module with $Z_3$ as its head) 
    from $\Delta$ and $\Upsilon$ (factors related to the outcome $Y$; captured by the left module with $Z_7$ as its head).
    The left module, itself, consists of a Parallel model 
    that attempts to proceed one step further and decompose $\Delta$ from $\Upsilon$.
% }
This is done with the help of a discrepancy penalty (see Section~\ref{sec:disc}).
% Figure~\ref{fig:combo} illustrates our designed belief net for the Hybrid model.

The decoder and encoder components of the Hybrid model 
--- $p(\cdot)$ and $q(\cdot)$ parametrized by $\theta_h$ and $\phi_h$ respectively --- 
involve the following distributions:
\begin{table}[H]
\setlength{\tabcolsep}{5pt}
%\vskip -.2in 
    \begin{center}
        \begin{tabular}{ll|l|ll}
        \multicolumn{2}{c|}{\bf Priors}  &\multicolumn{1}{c|}{\bf Likelihood}  &\multicolumn{2}{c}{\bf Posteriors}    \\
        \hline
        $p(z_2)$ & $p(z_1 | y, z_2)$  & $p(x | z_3, z_7)$ & $q(z_1 | z_7)$    & $q(y | z_1, z_5)$   \\
        $p(z_4)$ & $p(z_3 | t, z_4)$  &                   & $q(z_2 | y, z_1)$ & $q(t | z_3)$        \\
        $p(z_6)$ & $p(z_5 | y, z_6)$  &                   & $q(z_3 | x, y)$   &\\
                 & $p(z_7 | z_1, z_5)$&                   & $q(z_4 | t, z_3)$ &\\
                 &                    &                   & $q(z_5 | z_7)$    &\\
                 &                    &                   & $q(z_6 | y, z_5)$ &\\
                 &                    &                   & $q(z_7 | x, t)$   &\\
        \end{tabular}
    \end{center}
%\vskip -.3in 
\end{table}

Here, the conditional log-likelihood can be upper bounded by:
\begin{align}
    \sum_{i=1}^{N} \log p(x_i | t_i, y_i) \quad \geq \quad
    \sum_{i=1}^{N} \mathbb{E}_{q(z_3, z_7 | x, t ,y)} \big[ \log p(x_i |z_{3_i}, z_{7_i}) \big] \label{eq:recl_h} \\
    -\KL{q(z_1 | z_7)}{p(z_1 | y, z_2)}
    -\KL{q(z_2 | y, z_1)}{p(z_2)}          \nonumber \\
    -\KL{q(z_3 | x, y)}{p(z_3 | t, z_4)}
    -\KL{q(z_4 | t, z_3)}{p(z_4)}          \nonumber \\
    -\KL{q(z_5 | z_7)}{p(z_5 | y, z_6)}
    -\KL{q(z_6 | y, z_5)}{p(z_6)}          \nonumber \\
    -\KL{q(z_7 | x, t)}{p(z_7 | z_1, z_5)}  \label{eq:kld_h}
\end{align}

% \add[RG]
{For all three of these models,}
we refer to the first term in the ELBO 
(\ie right-hand-side of Equations~(\ref{eq:recl_s}),~(\ref{eq:recl_p}),~or~(\ref{eq:recl_h})) 
as % is called 
the Reconstruction Loss (RecL)
and the next term(s) 
(\ie Equations~(\ref{eq:kld_s}),~(\ref{eq:kld_p}),~or~(\ref{eq:kld_h})) 
% is referred to
as the KL 
    Divergence
(KLD).
Concisely, 
the ELBO can be 
viewed as maximizing: % written as:
$
\textrm{RecL} - \textrm{KLD}
$. %, which is to be maximized.
% \note[RG]{Given that Eq7 is MINIMIZING, why not write this in a similar form:  -RecL + KLD ??} \note[NH]{Since ELBO is a lower bound, we wish to maximize it. As a loss, we minimize its negation.}
 
\subsection{Further Disentanglement with {$\beta$-VAE}}
\label{sec:beta}
As mentioned earlier, we want the learned latent variables to be disentangled,
to match our assumption of non-overlapping factors $\Gamma$, $\Delta$, and $\Upsilon$.
To encourage this, 
we employ the \hbox{$\beta$-VAE} \citep{higgins2017beta}, 
which adds a hyperparameter $\beta$ 
% (usually greater than $1$) 
as a multiplier of the KLD part of the ELBO.
This adjustable hyperparameter facilitates a \mbox{trade-off} that helps balance the latent channel capacity and independence constraints 
% (handled by the KL terms) 
with the reconstruction accuracy
--- \ie including the $\beta$ hyperarameter should grant a better control over the level of disentanglement in the learned representations~\citep{burgess2018understanding}.
Therefore, the generative objective to be minimized becomes:
\begin{align}
\begin{split}
    \mathcal{L}_{\textrm{VAE}} \quad = \quad
    - \textrm{RecL} \;+\; \beta \cdot \textrm{KLD}
\end{split}
\end{align}
Although \citet{higgins2017beta} suggest 
setting $\beta$ % the $\beta$ to be set 
greater than $1$ in most applications,
\citet{hoffman2017the} show that having a $\beta<1$ weight on the KLD term can be interpreted as 
optimizing the ELBO under an alternative prior, 
which functions as a regularization term to 
reduce the chance of % prevent
degeneracy.

\subsection{Discrepancy}
\label{sec:disc}
Although all three proposed models encourage statistical independence between $T$ and $Z_1$ 
in the marginal posterior $q_{\phi}(Z_1 | T)$
% \add[RG]
{where $X$ is not given}
--- see the collider structure (at $X$): $T~\rightarrow~X~\leftarrow~Z_1$ in Figure~\ref{fig:series}
% note[RG]{But isn't X observed ... which means they are NOT independent?}
% \note[NH]{in the marginal posterior $q_{\phi}(Z_1 | T)$, $X$ is not given.}
---
% and $T \rightarrow Z_3 \rightarrow X \leftarrow Z_1$ in Figure~\ref{fig:parallel}),
an information leak is quite possible 
due to the correlation between the outcome~$Y$ and treatment~$T$ in the data.
We therefore require an extra regularization term on $q_{\phi}(Z_1 | T)$ 
in order to penalize the discrepancy (denoted by $\mathtt{disc}$) 
between the conditional distributions of $Z_1$ given $T\!=\!0$ versus given $T\!=\!1$.%
\footnote{
    Note that even for the Hybrid model (see Figure~\ref{fig:combo}),
    we apply the $\mathtt{disc}$ penalty only on $Z_1$ and not $Z_7$.
    This is because we want $Z_1$ to capture $\Upsilon$
    and $Z_5$ to capture $\Delta$ (so $Z_5$ should have a non-zero $\mathtt{disc}$). 
    Hence, 
    $Z_7$ must include both $\Delta$ and $\Upsilon$ (and therefore, it should have a non-zero $\mathtt{disc}$) to be able to reconstruct $X$.
}
To achieve this regularization, 
we calculate the $\mathtt{disc}$ using an Integral Probability Metric (IPM) \citep{mansour2009domain}
\footnote{
In this work, we use the Maximum Mean Discrepancy (MMD) \citep{gretton2012kernel} as our IPM.
}
(\cf \citep{louizos2015variational,shalit17a,yao2018representation}, etc.)
that measures the distance between the two above-mentioned distributions:
\begin{equation}
    \mathcal{L}_{\mathtt{disc}} 
    \quad = \quad 
    \textrm{IPM}\big(\ \{ z_1\}_{i:t_i=0}, \ \{ z_1 \}_{i:t_i=1} \ \big)
\end{equation}

\subsection{Predictive Loss}
Note, however, that neither the VAE nor the $\mathtt{disc}$ losses contribute to training a predictive model for outcomes.
To remedy this, we extend the objective function to include a discriminative term for the regression loss of predicting $y$:
\footnote{
    This is similar to the way \citet{kingma2014semi} 
    included a classification loss in their Equation~(9).
}
\begin{equation}
    \mathcal{L}_{\textrm{pred}} \quad = \quad
    \frac{1}{N} \sum_{i=1}^{N} \omega_i \cdot
    \mathcal{L}\big[\, y_i,\, \hat{y}_i \big]
\end{equation}
where 
the predicted outcome $\hat{y}_i = \mathbb{E}\big[\, q_{\phi}^{t_i}(y_i | z_{1_i}) \,\big]$; 
% is derived as the mean of the $q_{\phi}^{t_i}(y_i | z_{1_i})$ posterior trained for the respective treatment~$t_i$;
$\mathcal{L}\big[\, y_i,\, \hat{y}_i \,\big]$ is the factual loss 
(\ie L2 loss for real-valued outcomes and log loss for binary-valued outcomes); 
and
$\omega_i$ represent the weights 
% \add[R1]{
    that attempt to account for selection bias.
    We consider two approaches in the literature to derive the weights:
    (i) the \emph{Population-Based}~(PB) weights as proposed in {\citep{shalit17a}}; and 
    (ii) the \emph{Context-Aware}~(CA) weights as proposed in {\citep{hassanpour2019counterfactual}}.
    Note that disentangling $\Delta$ from $\Upsilon$ is only beneficial when using the CA weights,
    since we need just the $\Delta$ factors to derive them {\citep{hassanpour2020learning}}.
% }

\subsection{Final Model(s)}
\label{sec:final}
Putting everything together, the overall objective function to be minimized is:
\begin{equation}
    \mathcal{J} \quad = \quad 
    \mathcal{L}_{\textrm{pred}} \;+\;
    \alpha \cdot \mathcal{L}_{\mathtt{disc}} \;+\;
    \gamma \cdot \mathcal{L}_{\textrm{VAE}} \;+\;
    \lambda \cdot \mathfrak{Reg}
\end{equation}
where $\mathfrak{Reg}$ penalizes the model complexity. %

This objective function is motivated by the work of \citep{mccallum2006multi},
which suggested optimizing a convex combination of discriminative and generative losses would indeed improve predictive performance.
As an empirical verification, note that for $\gamma\!=\!0$, the Series and Parallel models effectively reduce to \hbox{CFR-Net}.
However, our empirical results (see Section~\ref{sec:exp}) suggest that
the generative term in the objective function helps learning representations that embed more relevant information for estimating outcomes than that of $\Phi$ in \hbox{CFR-Net}.

We refer to the family of our proposed methods as {VAE-CI} 
(Variational Auto-Encoder for Causal Inference);
specifically: \textbf{\hbox{\{S, P, H\}-VAE-CI}}, 
for \textbf{S}eries, \textbf{P}arallel, and \textbf{H}ybrid respectively.
We anticipate that each method is an \emph{improvement} over the previous one in terms of estimating causal effects,
culminating in \textbf{H-VAE-CI},
which we expect can best decompose the underlying factors and 
accurately estimate the outcomes of all treatments.

\section{Experiments, Results, and Discussion}
\label{sec:exp}
\subsection{Benchmarks}
\label{sec:bench}
% {Researchers often evaluate treatment effect estimation on}
% semi- or fully- synthetic datasets that include both factual and counterfactual outcomes.

\begin{enumerate}%[itemsep=0pt, topsep=0pt, leftmargin=*]
    \item 
\textbf{Infant Health and Development Program (IHDP)} $\quad$
The original IHDP randomized controlled trial was designed to evaluate the effect of specialist home visits on future cognitive test scores of premature infants.
Hill~\citep{hill2011bayesian} induced selection bias by removing a non-random subset of the treated population.
The dataset contains 747 instances (608 control and 139 treated) 
with $25$ covariates.
We use the same benchmark (with $100$ realizations of outcomes)
provided by and used in \citep{johansson2016learning} and \citep{shalit17a}. 

    \item 
\textbf{Atlantic Causal Inference Conference 2018 (ACIC'18)} $\quad$
ACIC'18 is a collection of binary-treatment observational datasets released for a data challenge.
Following \citep{shi2019adapting}, 
we used those datasets with $N \! \in \! \{1, 5, 10\} \! \times \! 10^3$ instances
(four datasets in each category).
The covariates matrix for each dataset involves 177 features 
and is sub-sampled from a table of medical measurements 
taken from the Linked Birth and Infant Death Data (LBIDD)~\citep{macdorman1998infant},
that contains information corresponding to 100,000 subjects.

    \item 
\textbf{Fully Synthetic Datasets} $\quad$
We generated a set of synthetic datasets according to the procedure described in \citep{hassanpour2020learning} 
(details in Appendix~\ref{app:syn}).
We considered all the viable datasets in a mesh generated by various sets of variables, 
of sizes $m_\Gamma, m_\Delta, m_\Upsilon \in \{0, 4, 8\}$ and $m_\Xi\!=\!1$.
This creates $24$ scenarios%
\footnote{
    There are $3^3\!=\!27$ combinations in total;
    however, we removed three of them that generate pure noise outcomes
    --- \ie $\Delta\!=\!\Upsilon\!=\!\emptyset$: $(0,0,0)$, $(4,0,0)$, and $(8,0,0)$.
}
that consider all relevant combinations of sizes of $\Gamma$, $\Delta$, and $\Upsilon$
(corresponding to different levels of selection bias).
For each scenario,
we synthesized multiple datasets with various initial random seeds to allow for statistical significance testing of the performance comparisons between the contending methods.
\end{enumerate}

\subsection{Identification of the Underlying Factors}
\subsubsection{Procedure for Evaluating Identification of the Underlying Factors} 
\begin{figure}[t]
    \begin{center}
    \includegraphics[width=.7\columnwidth]{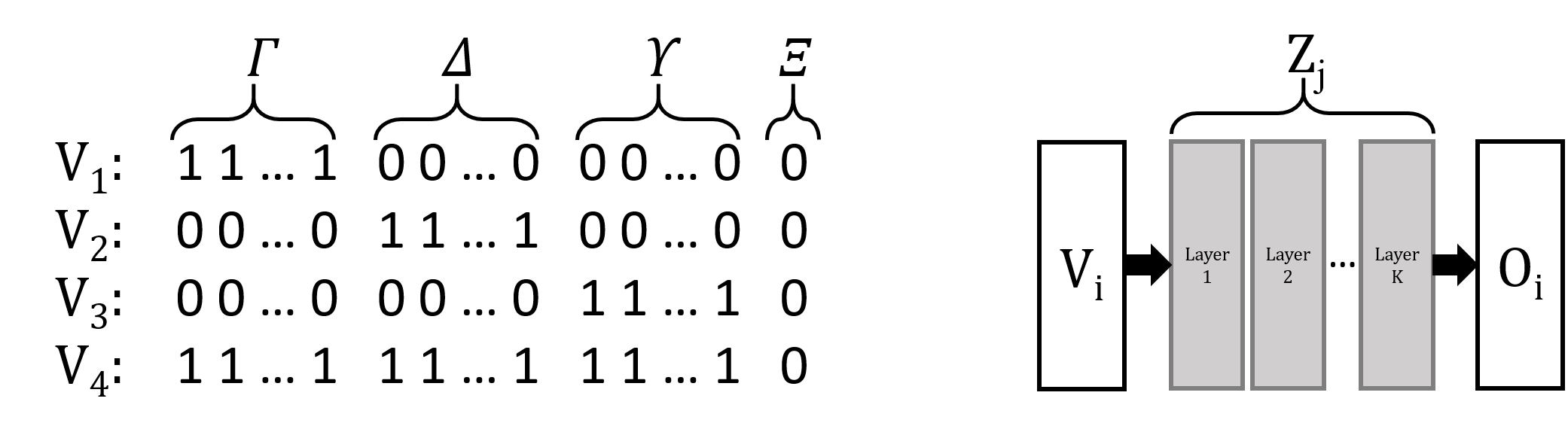}
    \end{center}
% %%\vskip -.1in
    \caption{The four dummy $x$-like vectors (left); and the input/output vectors of the representation networks (right).}
\label{fig:dummy}
\end{figure}

\quad
\label{app:weights}
To evaluate the identification performance of the underlying factors, 
we use a fully synthetic dataset (\#3 above) with $m_{\Gamma}\!=\!m_{\Delta}\!=\!m_{\Upsilon}\!=\!8$ and $m_\Xi\!=\!1$.
We 
% \change[RG]{produced four dummy vectors}
{then ran the learned model on four dummy test instances}
$V_i \in \Real^{m_\Gamma+m_\Delta+m_\Upsilon+m_\Xi}$ as depicted on the left-side of Figure~\ref{fig:dummy}.
% \note{So the training set had only 4 instances?  I am not tracking...} \note[NH]{This is actually the test set ... that measures how much information is able to pass through the neural network channel.}
The first to third vectors had ``1'' (constant) in the positions associated with $\Gamma$, $\Delta$, and $\Upsilon$ respectively,
and the remaining 17 positions were filled with ``0''.
The fourth vector was all ``1'' except for the last position (the noise) which was ``0''.
This helps measure the maximum amount of information that is passed to the final layer of each representation network.%

Next, each vector $V_i$ is fed to each trained network $Z_j$ (as if it was $x$).
We let $O_{i,j}$ be the $\mathtt{relu}$ output (here, $\in\!\Real^{200}$) of the encoder network $Z_j$ when $x\!=\!V_i$.
The average of the $200$ values of the $O_{i,j}$ (\ie $Avg(O_{i,j})$) 
represents the power of signal that was produced by the $Z_j$ channel on the input $V_i$.
% \note{Each instance is a 200-tuples. But there are only 4 of them?} \note[NH]{Yes, 200 is the number of learned features.}
The values reported in the tables illustrated in Figure~\ref{fig:weight} are the ratios of $Avg(O_{1,j})$, $Avg(O_{2,j})$, and $Avg(O_{3,j})$ divided by $Avg(O_{4,j})$ for each of the learned representation networks.
Note that, 
a larger ratio indicates that the respective representation network $Z_j$ has allowed more of the input signal $V_i$ to pass through.%
\footnote{
    Unlike the evaluation strategy presented in {\citep{hassanpour2020learning}} 
    that only examined the first layer's weights of each representation network, 
    we propagate the values through the entire network and check how much of each factor is exhibited in the final layer of every representation network.
    % Yet, the proposed procedure still crudely evaluates
    % the quality of disentanglement of the underlying factors in observational studies.
    % We did explore using the Mutual Information {\citep{belghazi2018mine}} for this task
    % (not shown here);
    % however, it appears that it does not work for high-dimensional data such as ours.
    % All in all, more research is needed to address this task.
}

% \add[RG]
{If the model could perfectly learn each underlying factor in a disentangled manner, 
we expect to see one element in each column to be significantly larger than the other elements in that column.
For example, for \hbox{H-VAE-CI},
the weights of the $\Gamma$ row should be highest for $Z_3$, as that means $Z_3$ has captured the $\Gamma$ factors.
Similarly, we would want the $Z_5$ and $Z_1$ entries on the $\Delta$ and $\Upsilon$ rows to be largest respectively.}

\comment{
    Figure~\ref{fig:weight} represents the relative (ratio) exhibition of the first three dummy vectors 
    (representing each underlying factor) 
    with respect to the fourth dummy vector,
    at the output layer of every representation network.
}
\subsubsection{Results' Analysis} 
\begin{figure*}[t]
%%\vskip -.75cm
\centering
    \includegraphics[width=.95\columnwidth]{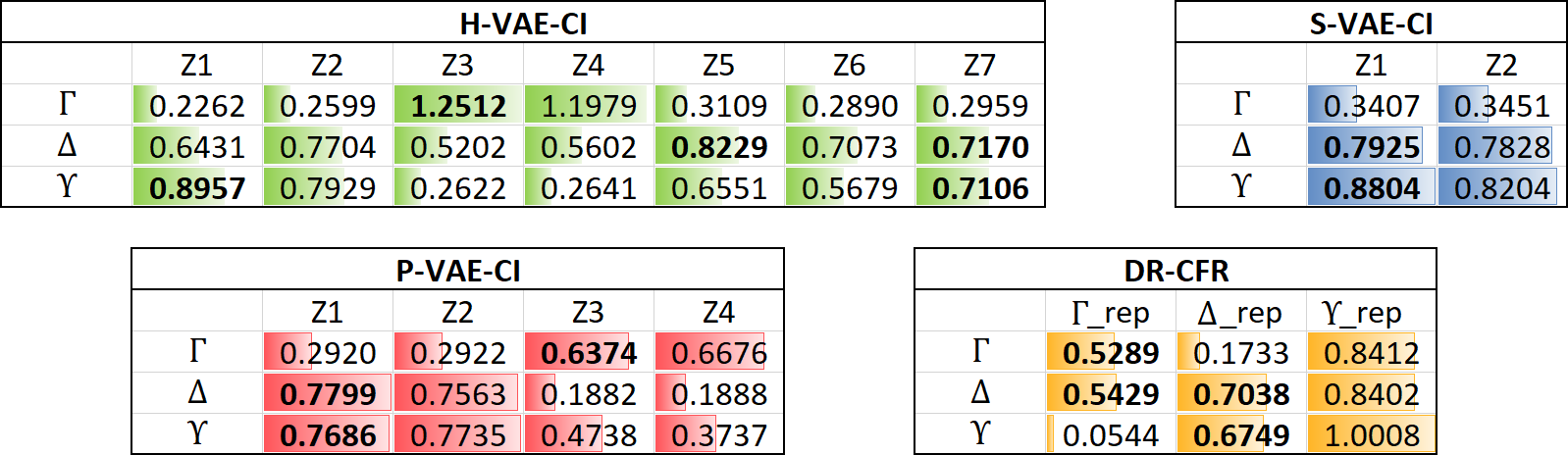}
    \caption{
        Performance analysis for decomposition of the underlying factors on the \textbf{Synthetic} dataset with $m_{\Gamma}\!=\!m_{\Delta}\!=\!m_{\Upsilon}\!=\!8$ and $m_\Xi\!=\!1$.
        % \add[RG]
        {(The color shading in each cell represents the value of that cell, with a larger amount of color for larger values.)}
        % The colors correspond to the legend in Figure~\ref{fig:pehe}.
        % The analysis is done according to the representation networks' weights.
    }
\label{fig:weight}
\end{figure*}

\quad
As expected, Figure~\ref{fig:weight} shows that
$Z_3$ and $Z_4$ capture $\Gamma$ (\eg the $Z_3$ ratios for $\Gamma$ in the {\{P, H\}-VAE-CI} tables are largest), 
and $Z_1$, $Z_2$, $Z_5$, $Z_6$, and $Z_7$ capture $\Delta$ and $\Upsilon$.
Note that decomposition of $\Delta$ from $\Upsilon$ has not been achieved by any of the methods 
% \footnote{
%     We do not expect this to decrease the ITE performance though, 
%     since both $\Delta$ and $\Upsilon$ contribute to outcome.
% }
except for {H-VAE-CI}, 
which captures $\Upsilon$~by~$Z_1$
and $\Delta$~by~$Z_5$
% \add[R4]{
    (note the ratios are largest for $Z_1$ and $Z_5$).
% }
This decomposition is vital for deriving context-aware importance sampling weights
because they must be calculated from $\Delta$ only \citep{hassanpour2020learning}.
Also observe that {\{P, H\}-VAE-CI} are each able to separate $\Gamma$ from $\Delta$.
However, {DR-CFR},
which tried to disentangle all factors, 
failed not only to disentangle $\Delta$ from $\Upsilon$,
but also $\Gamma$ from $\Delta$.

\subsection{Treatment Effect Estimation}
\label{sec:tee}
% \change[RG]{
% Evaluation of treatment effect estimation is often done with}
There are two categories of performance measures:

\noindent\textbf{Individual-based:}
    ``Precision in Estimation of Heterogeneous Effect'' \citep{hill2011bayesian}:
\begin{equation}
    \textrm{{\small PEHE}}
    \!=\!
    \sqrt{ \frac{1}{N} \sum_{i=1}^{N} \left( \hat{\textrm{e}}_i \!-\! \textrm{e}_i \right)^2 }
\end{equation}
uses $\hat{\textrm{e}}_i=\hat{y}_i^1 - \hat{y}_i^0$ as the estimated effect and $\textrm{e}_i=y_i^1 - y_i^0$ as the true effect.

\noindent\textbf{Population-based:}
    ``Bias of the Average Treatment Effect'':
\begin{equation}
    \epsilon_{\textrm{ATE}}
    \!=\!
    \big|\textrm{ATE} - \widehat{\textrm{ATE}}\big|
\end{equation}
% \note[RG]{Are there $N$ instances with t=0 and $N$ with t=1 ?} \note[NH]{yes, because here we know the counterfactuals. so we know both $y^0$ and $y^1$ for all instances.}
where 
$\, \textrm{ATE}=\frac{1}{N} \sum_{i=1}^{N} y^1_i-\frac{1}{N} \sum_{j=1}^{N} y^0_j \,$ 
% in which $y^1_i$ and $y^0_j$ are the true outcomes %for the respective treatments
and $\widehat{\textrm{ATE}}$ follows the same formula except that it is calculated based on the estimated outcomes.

In this paper, we compare performances of the proposed \textbf{\hbox{\{S, P, H\}-VAE-CI}} 
versus the following treatment effect estimation methods:
\textbf{\hbox{CFR-Net}}~\citep{shalit17a},
\textbf{\hbox{DR-CFR}}~\citep{hassanpour2020learning},
\textbf{\hbox{Dragon-Net}}~\citep{shi2019adapting},
\textbf{GANITE}~\citep{yoon2018ganite}, 
\textbf{CEVAE}~\citep{louizos2017causal}, and
\textbf{TEDVAE}~\citep{zhang2021treatment}.
The basic search grid for hyperparameters of the \hbox{CFR-Net} based algorithms (including our methods) is available in Appendix~\ref{app:grid}.
% For the other algorithms, 
% we searched 
% around
% their default hyperparameter settings.
% \footnote{
    To ensure that our performance gain is not merely based on an increased complexity of the models, 
    we also performed our grid search for all the contending methods with an updated number of layers and/or number of neurons in each layer,
    This guaranteed that all methods have the same number of parameters and therefore
    a similar model complexity. 
% }

We ran the experiments for the contender methods using their publicly available code-bases;
note the following points regarding these runs: % though
\begin{itemize}%[itemsep=0pt, topsep=0pt, leftmargin=*]
    \item   Since Dragon-Net is designed to estimate ATE only, 
            we did not report its performance results for the {\small PEHE} measure 
            (which, as expected, were significantly inaccurate). % compared to the rest of the methods).
    \item   Original GANITE code-base 
    % \change{as implemented for binary outcomes only}
    {could only deal with binary outcomes}.
            We modified the code (losses, etc.) 
            to allow it to % [][] such that it could
            process real-valued outcomes also.
    \item   We were surprised that CEVAE diverged when running on the ACIC'18 datasets.
            To avoid this, we had to run the ACIC'18 experiments on the binary covariates only.
\end{itemize}

\subsubsection{Results' Analysis} \quad
Table~\ref{tab:perf} summarizes the mean and standard deviation of the {\small PEHE} and $\epsilon_{\textrm{ATE}}$ measures (lower is better) on the IHDP, ACIC'18, and Synthetic benchmarks.
{VAE-CI} achieves the best performance among the contending methods.
These results are statistically significant 
(in \textbf{bold}; based on the Welch's unpaired t-test with $\alpha\!=\!0.05$)
for the IHDP and Synthetic benchmarks.
Although {VAE-CI} also achieves the best performance on the ACIC'18 benchmark, 
the results are not statistically significant due to the high standard deviation of the performances of the contending methods.

\begin{table*}[t]
\caption{{\small PEHE} and $\epsilon_{\textrm{ATE}}$ performance measures (lower is better) represented in the form of ``mean {\scriptsize (standard deviation)}''.
% \add[RG]
{The {\bf bold} values in each column are the best (statistically significant based on the Welch's unpaired t-test with $\alpha\!=\!0.05$).}
}
\label{tab:perf}
\centering
\begin{tabular}{lcccccc}
\toprule
    \multirow{2}{*}{\bfseries Method} & 
    \multicolumn{2}{c}{\bfseries IHDP} &
    \multicolumn{2}{c}{\bfseries ACIC'18} &
    \multicolumn{2}{c}{\bfseries Synthetic} \\ &
    \textbf{\small{PEHE}} & \pmb{$\epsilon_{\textrm{ATE}}$} & 
    \textbf{\small{PEHE}} & \pmb{$\epsilon_{\textrm{ATE}}$} & 
    \textbf{\small{PEHE}} & \pmb{$\epsilon_{\textrm{ATE}}$} \\
\midrule
    \textbf{CFR-Net}    &	 0.75 {\scriptsize (0.57)}    	&	 0.08 {\scriptsize (0.10)}	&	 5.13 {\scriptsize (5.59)}    	&	 1.21 {\scriptsize (1.81)}	&	 0.39 {\scriptsize (0.08)}    	&	 0.027 {\scriptsize (0.020)}	\\
    \textbf{DR-CFR}  	&	 0.65 {\scriptsize (0.37)}    	&	 0.03 {\scriptsize (0.04)}	&	 3.86 {\scriptsize (3.39)}    	&	 0.80 {\scriptsize (1.41)}	&	 0.26 {\scriptsize (0.07)}    	&	 0.007 {\scriptsize (0.004)}	\\
    \textbf{Dragon-Net} &	 NA                           	&	 0.14 {\scriptsize (0.15)}	&	 NA                           	&	 0.48 {\scriptsize (0.77)}	&	 NA                           	&	 0.007 {\scriptsize (0.005)}	\\
    \textbf{GANITE}  	&	 2.81 {\scriptsize (2.30)}    	&	 0.24 {\scriptsize (0.46)}	&	 3.55 {\scriptsize (2.27)}    	&	 0.69 {\scriptsize (0.65)}	&	 1.28 {\scriptsize (0.43)}    	&	 0.036 {\scriptsize (0.015)}	\\
    \textbf{CEVAE}   	&	 2.50 {\scriptsize (3.47)}    	&	 0.18 {\scriptsize (0.25)}	&	 5.30 {\scriptsize (5.52)}    	&	 3.29 {\scriptsize (3.50)}	&	 1.39 {\scriptsize (0.32)}    	&	 0.287 {\scriptsize (0.217)}	\\
    \textbf{TEDVAE}   	&	 1.61 {\scriptsize (2.37)}    	&	 0.18 {\scriptsize (0.23)}	&	 6.63 {\scriptsize (8.69)}      &	 3.74 {\scriptsize (5.00)}  &	 0.25 {\scriptsize (0.07)}    	&	 0.013 {\scriptsize (0.007)}	\\
\midrule
    \textbf{\hbox{S-VAE-CI}}  	&	 \textbf{0.51 {\scriptsize (0.37)}}  	&	 \textbf{0.00 {\scriptsize (0.02)}}	&	 2.73 {\scriptsize (2.39)}  	&	 0.51 {\scriptsize (0.82)}	&	 0.28 {\scriptsize (0.05)}  	&	 0.004 {\scriptsize (0.003)}	\\
    \textbf{\hbox{P-VAE-CI}}  	&	 \textbf{0.52 {\scriptsize (0.36)}}  	&	 \textbf{0.01 {\scriptsize (0.03)}}	&	 2.62 {\scriptsize (2.26)}  	&	 0.37 {\scriptsize (0.75)}	&	 0.28 {\scriptsize (0.05)}  	&	 0.004 {\scriptsize (0.003)}	\\
    \textbf{\hbox{H-VAE-CI (PB)}}  	&	 \textbf{0.49 {\scriptsize (0.36)}}  	&	 \textbf{0.01 {\scriptsize (0.02)}}	&	 1.78 {\scriptsize (1.27)}  	&	 0.44 {\scriptsize (0.77)}	&	 \textbf{0.20 {\scriptsize (0.03)}}  	&	 \textbf{0.003 {\scriptsize (0.002)}}	\\
    \textbf{\hbox{H-VAE-CI (CA)}}  	&	 \textbf{0.48 {\scriptsize (0.35)}}  	&	 \textbf{0.01 {\scriptsize (0.01)}}	&	 1.66 {\scriptsize (1.30)}  	&	 0.39 {\scriptsize (0.75)}	&	 \textbf{0.18 {\scriptsize (0.02)}}  	&	 \textbf{0.003 {\scriptsize (0.002)}}	\\
\bottomrule
\end{tabular}
\end{table*}

\begin{figure}[t]
\centering
    \includegraphics[width=.75\columnwidth]{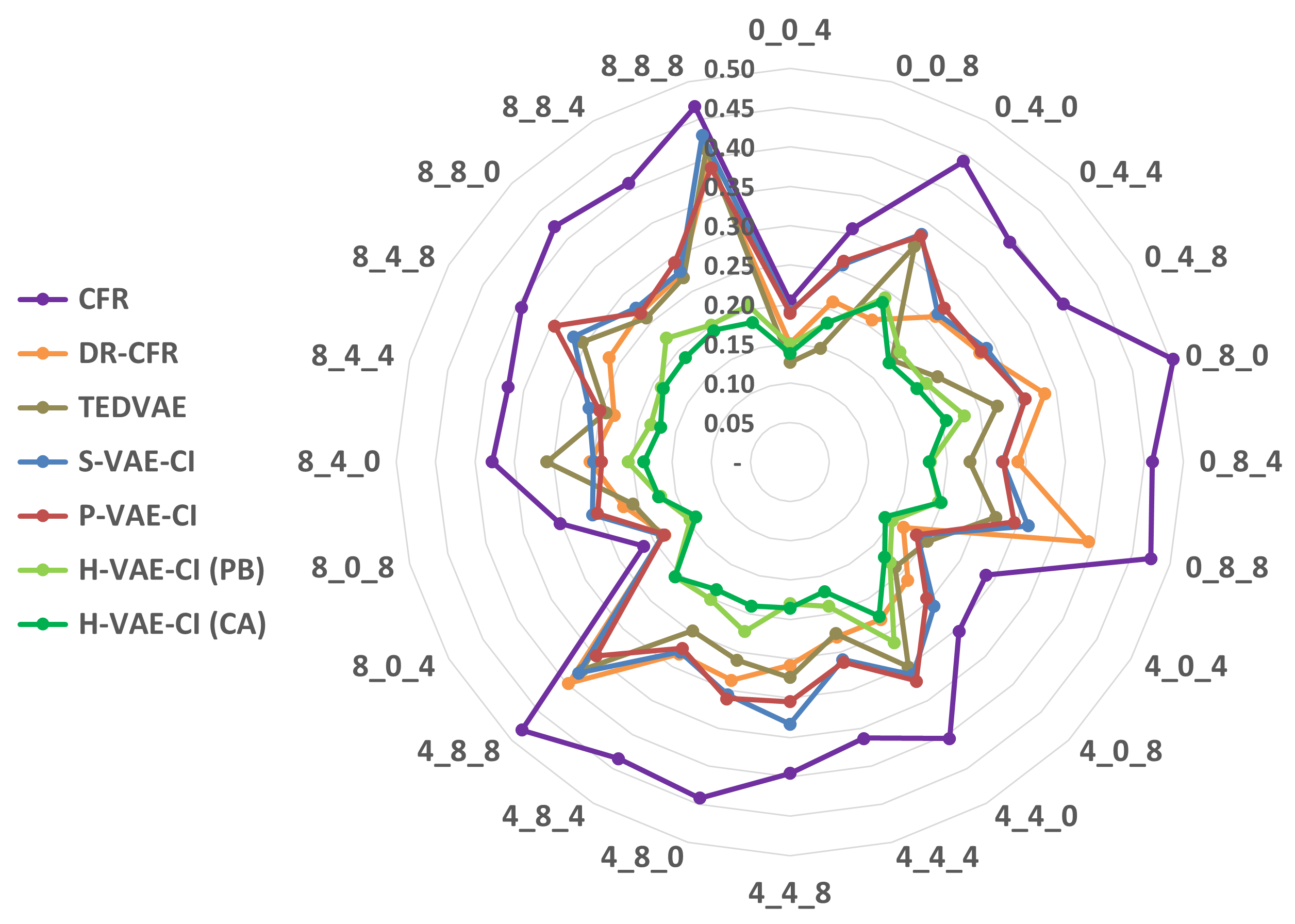} %}
    \caption{Radar graphs of {\small PEHE} (on the radii; closer to the center is better) for the entire \textbf{Synthetic} benchmark ($24 \times 3$ with $N\!=\!10,\!000$; each vertex denotes the respective dataset). Figure is best viewed in color.
    }
\label{fig:pehe}
\end{figure}

Figure~\ref{fig:pehe} visualizes the {\small PEHE} measures on the entire synthetic datasets with sample size of $N\!=\!10,\!000$.
We observe that both plots corresponding to \hbox{H-VAE-CI} method (PB as well as CA) are 
% \change{inscribed by}
{completely within the} 
plots of all other methods,
showcasing \hbox{H-VAE-CI}'s superior performance under every possible selection bias scenario.
% \add[R1]{
    Note that for scenarios where $m_{\Delta}\!=0$ 
    ({\ie} the ones of the form $m_\Gamma$\_$0$\_$m_\Upsilon$ on 
    the % [][]
    perimeter of the radar chart in {Figure~\ref{fig:pehe}}),
    % 0\_0\_4, 0\_0\_8, 4\_0\_4, 4\_0\_8, 8\_0\_4, and 8\_0\_8
    the performances of \hbox{H-VAE-CI~(PB)} and \hbox{H-VAE-CI (CA)} are almost identical.
    This is expected,
    since for these scenarios, 
    the learned representation for $\Delta$ would be degenerate,
    and therefore, 
    the context-aware weights would reduce to population-based ones.
    On the other hand, 
    for scenarios where $m_{\Delta}\! \neq 0$,
    the \hbox{H-VAE-CI (CA)} often performs better than \hbox{H-VAE-CI (PB)}.
    This 
    % \change{can be attributed to the fact that}
    {may be because}
    \hbox{H-VAE-CI} has correctly disentangled $\Delta$ from $\Upsilon$.
    This facilitates learning good CA weights that better account for selection bias,
    which in turn, results in a better causal effect estimation performance.
% }

\subsection{Hyperparameters' Sensitivity Analyses}
\label{app:hyper}
Figure~\ref{fig:hyper_analyses} illustrates
the results of our hyperparameters' sensitivity analyses 
(in terms of {\small PEHE}).
In the following, we discuss the insights we gained from these ablation studies:

\begin{figure*}[h]
    \begin{center}
        \subfigure[$\alpha$]{\label{fig:h_a}\includegraphics[width=.32\linewidth]{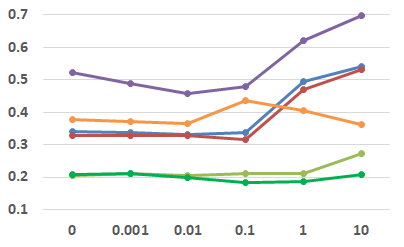}}
        \subfigure[$\beta$]{\label{fig:h_b}\includegraphics[width=.32\linewidth]{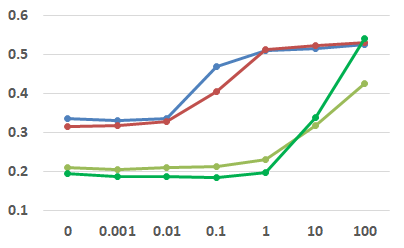}}
        \subfigure[$\gamma$]{\label{fig:h_g}\includegraphics[width=.32\linewidth]{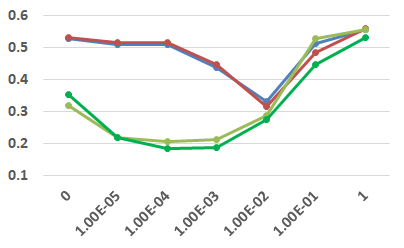}}
    \end{center}
% %%\vskip -.15in
    \caption{
        Hyperparameters' ($x$-axis) sensitivity analysis based on {\small PEHE} ($y$-axis) on the \textbf{synthetic} dataset with $m_{\Gamma, \Delta, \Upsilon}\!=\!8, m_\Xi\!=\!1$.
        Legend is the same as Figure~\ref{fig:pehe}:
        purple for \hbox{CFR-Net}, orange for \hbox{DR-CFR}, blue for \hbox{S-VAE-CI}, red for \hbox{P-VAE-CI}, and light and dark green for \hbox{H-VAE-CI} (PB) and (CA) respectively.
        % \note[RG]{Why not just repeat what the colors mean?}
        Plots are best viewed in color. 
    }
    \label{fig:hyper_analyses}
\end{figure*}

\begin{itemize}%[itemsep=0pt, topsep=0pt, leftmargin=*]
\item 
    For the $\alpha$ hyperparameter 
    (\ie coefficient of the discrepancy penalty),
    Figure~\ref{fig:h_a} suggests that {DR-CFR} and {H-VAE-CI} methods have the most robust performance
    % \add[R1 \& R4]{
        throughout various values of $\alpha$. 
        This is expected, because,
        unlike CFR-Net and \hbox{\{S, P\}-VAE-CI},
        \hbox{DR-CFR} and \hbox{H-VAE-CI} possess an independent node for representing $\Delta$.
        This helps them still capture $\Delta$ as $\alpha$ grows;
        since for them, $\alpha$ only affects learning a representation of $\Upsilon$.
        Comparing \hbox{H-VAE-CI (PB) with (CA)}, 
        we observe that %there exist values for 
        for all 
        $\alpha \!>\! 0.01$, 
        {(CA)} outperforms {(PB)}.
        This is because the discrepancy penalty would force 
        $Z_1$ to only capture $\Upsilon$ and $Z_5$ to only capture $\Delta$.
        This results in deriving better CA weights
        (that should be learned from $\Delta$; 
        here, from its learned representation~$Z_5$).
        {H-VAE-CI (PB)}, on the other hand, cannot take advantage of this disentanglement,
        which explains its sub-optimal performance.
    % }

\item 
    Figure~\ref{fig:h_b} shows that
    various $\beta$ values 
    (\ie coefficient of KL divergence penalty)
    % \add[R1 \& R3 \& R4]{
        do not make much difference for {H-VAE-CI} 
        (except for $\beta \geq 1$, 
        since this large value means 
        % which seems to be too strong,         enforcing
        the learned representations will be close to Gaussian noise).
        We initially thought % Our initial hypothesis in
        using {$\beta$-VAE} might help \emph{further} disentangle the underlying factors.
        However, {Figure~\ref{fig:h_b}} suggests that close-to-zero or even zero $\beta$s also work effectively.
        Our hypothesis is that the \hbox{H-VAE-CI}'s architecture already takes care of decomposing the $\Gamma$, $\Delta$, and $\Upsilon$ factors, without needing the help of a KLD penalty. 
        {Appendix~\ref{app:beta0}} includes more evidence and a detailed discussion on why this interpretation should hold.
    \comment{        
        In order to validate this hypothesis, 
        we examined the decomposition tables of \hbox{H-VAE-CI} 
        (similar to the performance reported in the green table in {Figure~\ref{fig:weight}}
        in Appendix~\ref{app:beta0}) 
        for extreme configurations with $\beta=0$ 
        and observed that they were all effective at 
        decomposing the underlying factors $\Gamma$, $\Delta$, and $\Upsilon$. 
        {Figure~\ref{fig:beta0}} shows several of these tables.
        This means either of the following is happening:
        (i)~\hbox{$\beta$-VAE} is not the best performing disentangling method and other disentangling constraints should be used instead 
            --- {\eg} works of {\citet{chen2018isolating}} and {\citet{lopez2018information}};
            or
        (ii)~it is theoretically impossible to achieve disentanglement without some supervision {\citep{locatello2019challenging}}, 
            which might not be possible to provide in this task.
        Exploring these options is out of the scope of this paper and is left to future work.
    }
\item 
    % \add[R1 \& R4]{
        For hyperparameter $\gamma$ 
        ({\ie} coefficient of the generative loss penalty),
        \hbox{H-VAE-CI} achieves the most stable performance compared to the \hbox{\{S, P\}-VAE-CI} models
        --- see {Figure~\ref{fig:h_g}}.
        % {H-VAE-CI} seems to be more resilient to various $\gamma$ values than the other two models.
\comment{
        Of particular interest is the superior performance of 
        \hbox{H-VAE-CI} for $\gamma \leq 0.01$ compared to that of \hbox{\{S, P\}-VAE-CI}. 
        This means that having the generative loss term ({\ie} $\mathcal{L}_{\textrm{VAE}}$) 
}
        Finding that
        \hbox{H-VAE-CI} for $\gamma \leq 0.01$ performs better than \hbox{\{S, P\}-VAE-CI},
        suggests that having the generative loss term ({\ie} $\mathcal{L}_{\textrm{VAE}}$) 
        is more important for \hbox{\{S, P\}-VAE-CI} than for \hbox{H-VAE-CI} to perform well
        --- note an extreme case happens at $\gamma=0$,
        where the latter performs %\annote
        {significantly}{ (statistically)}
        better than the former.
        We hypothesize that this is because
        \hbox{H-VAE-CI} already learns expressive representations $Z_3$ and $Z_7$, 
        meaning the optimization no~longer really \emph{{requires}} the $\mathcal{L}_{\textrm{VAE}}$ term to impose that. 
        This is in contrast to $Z_1$ in \hbox{S-VAE-CI}, 
        and $Z_1$ and $Z_3$ in \hbox{P-VAE-CI}.
    % } 
\end{itemize}

\section{Future works and Conclusion}
\label{sec:conc}
Despite the success of the proposed methods,
especially the Hybrid model, 
in addressing causal inference for treatment effect estimation, 
no known algorithms can yet learn to perfectly decompose factors $\Delta$ and $\Upsilon$.
% \add[R1]{
    This goal is important because 
    % we know 
    isolating $\Delta$,
    and learning Context-Aware (CA) weights from it,
    does enhance the quality of the causal effect estimation performance 
    --- note the superior performance of \hbox{H-VAE-CI (CA)}.
% }
% \add[R1 \& R3]{
    The results of our ablation study in {Figure~\ref{fig:h_b}}, however,
    revealed that the currently used \hbox{$\beta$-VAE} does not
    help much with disentanglement of the underlying factors.
      Therefore, the proposed architectures and objective function 
      ought to be responsible for most of the achieved decomposition.
    A future direction is to explore the use of better disentangling constraints 
    ({\eg} works of {\citep{chen2018isolating}} and {\citep{lopez2018information}})
    to see if that would yield % we can get 
    sharper results.
% }
% Although the current work proved to be a step in the right direction, 
% further future research is needed to perfect this disentanglement.
% This is left to future work.

The goal of this paper was to estimate causal effects 
(either for individuals or the entire population) 
from observational data.
We designed three models that employ Variational Auto-Encoders (VAE) \citep{kingma2014auto,rezende2014stochastic}, 
namely Series, Parallel, and Hybrid. 
Each model was an improvement over the previous one, in terms of 
identifying the underlying factors of any observational data as well as 
estimating the causal effects.
Our proposed methods employed \citet{kingma2014semi}'s M1 and M2 models as their building  blocks.
Our Hybrid model
performed best, and 
% , as the best performing architecture,
% partially 
succeeded at learning decomposed representations of the underlying factors; %$\Gamma$, $\Delta$, and $\Upsilon$;
this, in turn, helped to accurately estimate the outcomes of all treatments.
Our empirical results demonstrated the superiority of the proposed methods,
compared to both state-of-the-art discriminative as well as generative approaches in the literature. 

\bibliographystyle{abbrvnat}
\bibliography{refs}

\begin{thebibliography}{37}
\providecommand{\natexlab}[1]{#1}
\providecommand{\url}[1]{\texttt{#1}}
\expandafter\ifx\csname urlstyle\endcsname\relax
  \providecommand{\doi}[1]{doi: #1}\else
  \providecommand{\doi}{doi: \begingroup \urlstyle{rm}\Url}\fi

\bibitem[Burgess et~al.(2018)Burgess, Higgins, Pal, Matthey, Watters,
  Desjardins, and Lerchner]{burgess2018understanding}
C.~Burgess, I.~Higgins, A.~Pal, L.~Matthey, N.~Watters, G.~Desjardins, and
  A.~Lerchner.
\newblock Understanding disentangling in {$\beta$-VAE}.
\newblock \emph{arXiv preprint:1804.03599}, 2018.

\bibitem[Chen et~al.(2018)Chen, Li, Grosse, and Duvenaud]{chen2018isolating}
R.~T. Chen, X.~Li, R.~B. Grosse, and D.~K. Duvenaud.
\newblock Isolating sources of disentanglement in variational autoencoders.
\newblock In \emph{NeurIPS}, 2018.

\bibitem[Goodfellow et~al.(2014)Goodfellow, Pouget-Abadie, Mirza, Xu,
  Warde-Farley, Ozair, Courville, and Bengio]{goodfellow2014generative}
I.~Goodfellow, J.~Pouget-Abadie, M.~Mirza, B.~Xu, D.~Warde-Farley, S.~Ozair,
  A.~Courville, and Y.~Bengio.
\newblock Generative adversarial nets.
\newblock In \emph{NeurIPS}, 2014.

\bibitem[Gordon and Hern{\'a}ndez-Lobato(2020)]{gordon2020combining}
J.~Gordon and J.~M. Hern{\'a}ndez-Lobato.
\newblock Combining deep generative and discriminative models for bayesian
  semi-supervised learning.
\newblock \emph{Pattern Recognition}, 100, 2020.

\bibitem[Gretton et~al.(2012)Gretton, Borgwardt, Rasch, Sch{\"o}lkopf, and
  Smola]{gretton2012kernel}
A.~Gretton, K.~M. Borgwardt, M.~J. Rasch, B.~Sch{\"o}lkopf, and A.~Smola.
\newblock A kernel two-sample test.
\newblock \emph{JMLR}, 13\penalty0 (March), 2012.

\bibitem[Guo et~al.(2018)Guo, Cheng, Li, Hahn, and Liu]{guo2018survey}
R.~Guo, L.~Cheng, J.~Li, P.~R. Hahn, and H.~Liu.
\newblock A survey of learning causality with data: Problems and methods.
\newblock \emph{arXiv preprint:1809.09337}, 2018.

\bibitem[Harada and Kashima(2020)]{harada2020counterfactual}
S.~Harada and H.~Kashima.
\newblock Counterfactual propagation for semi-supervised individual treatment
  effect estimation.
\newblock \emph{ECML-PKDD}, 2020.

\bibitem[Hassanpour and Greiner(2019)]{hassanpour2019counterfactual}
N.~Hassanpour and R.~Greiner.
\newblock Counterfactual regression with importance sampling weights.
\newblock In \emph{IJCAI}, 2019.

\bibitem[Hassanpour and Greiner(2020)]{hassanpour2020learning}
N.~Hassanpour and R.~Greiner.
\newblock Learning disentangled representations for counterfactual regression.
\newblock In \emph{ICLR}, 2020.

\bibitem[Higgins et~al.(2017)Higgins, Matthey, Pal, Burgess, Glorot, Botvinick,
  Mohamed, and Lerchner]{higgins2017beta}
I.~Higgins, L.~Matthey, A.~Pal, C.~Burgess, X.~Glorot, M.~Botvinick,
  S.~Mohamed, and A.~Lerchner.
\newblock {$\beta$-VAE}: Learning basic visual concepts with a constrained
  variational framework.
\newblock \emph{ICLR}, 2017.

\bibitem[Hill(2011)]{hill2011bayesian}
J.~L. Hill.
\newblock Bayesian nonparametric modeling for causal inference.
\newblock \emph{Journal of Computational and Graphical Statistics}, 20\penalty0
  (1), 2011.

\bibitem[Hoffman et~al.(2017)Hoffman, Riquelme, and Johnson]{hoffman2017the}
M.~Hoffman, C.~Riquelme, and M.~Johnson.
\newblock The {$\beta$-VAE}'s implicit prior.
\newblock 2017.
\newblock URL \url{http://bayesiandeeplearning.org/2017/papers/66.pdf}.

\bibitem[Holland(1986)]{holland1986statistics}
P.~W. Holland.
\newblock Statistics and causal inference.
\newblock \emph{Journal of the American statistical Association}, 81\penalty0
  (396), 1986.

\bibitem[Imbens and Rubin(2015)]{imbens_rubin_2015}
G.~W. Imbens and D.~B. Rubin.
\newblock \emph{Causal Inference for Statistics, Social, and Biomedical
  Sciences: An Introduction}.
\newblock Cambridge University Press, 2015.

\bibitem[Johansson et~al.(2016)Johansson, Shalit, and
  Sontag]{johansson2016learning}
F.~Johansson, U.~Shalit, and D.~Sontag.
\newblock Learning representations for counterfactual inference.
\newblock In \emph{ICML}, 2016.

\bibitem[Kingma and Ba(2015)]{kingma2015adam}
D.~P. Kingma and J.~L. Ba.
\newblock Adam: A method for stochastic optimization.
\newblock In \emph{ICLR}, 2015.

\bibitem[Kingma and Welling(2014)]{kingma2014auto}
D.~P. Kingma and M.~Welling.
\newblock Auto-encoding variational bayes.
\newblock In \emph{ICLR}, 2014.

\bibitem[Kingma et~al.(2014)Kingma, Mohamed, Rezende, and
  Welling]{kingma2014semi}
D.~P. Kingma, S.~Mohamed, D.~J. Rezende, and M.~Welling.
\newblock Semi-supervised learning with deep generative models.
\newblock In \emph{NeurIPS}, 2014.

\bibitem[Kuang et~al.(2017)Kuang, Cui, Li, Jiang, Yang, and
  Wang]{kuang2017treatment}
K.~Kuang, P.~Cui, B.~Li, M.~Jiang, S.~Yang, and F.~Wang.
\newblock Treatment effect estimation with data-driven variable decomposition.
\newblock In \emph{AAAI}, 2017.

\bibitem[Locatello et~al.(2019)Locatello, Bauer, Lucic, Raetsch, Gelly,
  Sch{\"o}lkopf, and Bachem]{locatello2019challenging}
F.~Locatello, S.~Bauer, M.~Lucic, G.~Raetsch, S.~Gelly, B.~Sch{\"o}lkopf, and
  O.~Bachem.
\newblock Challenging common assumptions in the unsupervised learning of
  disentangled representations.
\newblock In \emph{ICML}, 2019.

\bibitem[Lopez et~al.(2018)Lopez, Regier, Jordan, and
  Yosef]{lopez2018information}
R.~Lopez, J.~Regier, M.~I. Jordan, and N.~Yosef.
\newblock Information constraints on auto-encoding variational bayes.
\newblock In \emph{NeurIPS}, 2018.

\bibitem[Louizos et~al.(2015)Louizos, Swersky, Li, Welling, and
  Zemel]{louizos2015variational}
C.~Louizos, K.~Swersky, Y.~Li, M.~Welling, and R.~Zemel.
\newblock The variational fair autoencoder.
\newblock \emph{arXiv preprint:1511.00830}, 2015.

\bibitem[Louizos et~al.(2017)Louizos, Shalit, Mooij, Sontag, Zemel, and
  Welling]{louizos2017causal}
C.~Louizos, U.~Shalit, J.~M. Mooij, D.~Sontag, R.~Zemel, and M.~Welling.
\newblock Causal effect inference with deep latent-variable models.
\newblock In \emph{NeurIPS}. 2017.

\bibitem[MacDorman and Atkinson(1998)]{macdorman1998infant}
M.~F. MacDorman and J.~O. Atkinson.
\newblock Infant mortality statistics from the 1996 period linked birth/infant
  death dataset.
\newblock \emph{Monthly Vital Statistics Report}, 46\penalty0 (12), 1998.

\bibitem[Mansour et~al.(2009)Mansour, Mohri, and
  Rostamizadeh]{mansour2009domain}
Y.~Mansour, M.~Mohri, and A.~Rostamizadeh.
\newblock Domain adaptation: Learning bounds and algorithms.
\newblock \emph{arXiv preprint:0902.3430}, 2009.

\bibitem[McCallum et~al.(2006)McCallum, Pal, Druck, and
  Wang]{mccallum2006multi}
A.~McCallum, C.~Pal, G.~Druck, and X.~Wang.
\newblock Multi-conditional learning: Generative/discriminative training for
  clustering and classification.
\newblock In \emph{AAAI}, 2006.

\bibitem[Ng and Jordan(2002)]{ng2002discriminative}
A.~Y. Ng and M.~I. Jordan.
\newblock On discriminative vs. generative classifiers: A comparison of
  logistic regression and naive bayes.
\newblock In \emph{NeurIPS}, 2002.

\bibitem[Pearl(2009)]{pearl2009causality}
J.~Pearl.
\newblock \emph{Causality}.
\newblock Cambridge University Press, 2009.

\bibitem[Peters et~al.(2017)Peters, Janzing, and
  Sch{\"o}lkopf]{peters2017elements}
J.~Peters, D.~Janzing, and B.~Sch{\"o}lkopf.
\newblock \emph{Elements of causal inference: foundations and learning
  algorithms}.
\newblock MIT press, 2017.

\bibitem[Rezende et~al.(2014)Rezende, Mohamed, and
  Wierstra]{rezende2014stochastic}
D.~J. Rezende, S.~Mohamed, and D.~Wierstra.
\newblock Stochastic backpropagation and approximate inference in deep
  generative models.
\newblock \emph{ICML}, 2014.

\bibitem[Rosenbaum and Rubin(1983)]{rosenbaum1983central}
P.~R. Rosenbaum and D.~B. Rubin.
\newblock The central role of the propensity score in observational studies for
  causal effects.
\newblock \emph{Biometrika}, 1983.

\bibitem[Rubin(1974)]{rubin1974estimating}
D.~B. Rubin.
\newblock Estimating causal effects of treatments in randomized and
  nonrandomized studies.
\newblock \emph{Journal of Educational Psychology}, 66\penalty0 (5), 1974.

\bibitem[Shalit et~al.(2017)Shalit, Johansson, and Sontag]{shalit17a}
U.~Shalit, F.~D. Johansson, and D.~Sontag.
\newblock Estimating individual treatment effect: Generalization bounds and
  algorithms.
\newblock In \emph{ICML}, 2017.

\bibitem[Shi et~al.(2019)Shi, Blei, and Veitch]{shi2019adapting}
C.~Shi, D.~Blei, and V.~Veitch.
\newblock Adapting neural networks for the estimation of treatment effects.
\newblock In \emph{NeurIPS}, 2019.

\bibitem[Yao et~al.(2018)Yao, Li, Li, Huai, Gao, and
  Zhang]{yao2018representation}
L.~Yao, S.~Li, Y.~Li, M.~Huai, J.~Gao, and A.~Zhang.
\newblock Representation learning for treatment effect estimation from
  observational data.
\newblock In \emph{NeurIPS}, 2018.

\bibitem[Yoon et~al.(2018)Yoon, Jordon, and van~der Schaar]{yoon2018ganite}
J.~Yoon, J.~Jordon, and M.~van~der Schaar.
\newblock {GANITE}: Estimation of individualized treatment effects using
  generative adversarial nets.
\newblock In \emph{ICLR}, 2018.

\bibitem[Zhang et~al.(2021)Zhang, Liu, and Li]{zhang2021treatment}
W.~Zhang, L.~Liu, and J.~Li.
\newblock Treatment effect estimation with disentangled latent factors.
\newblock In \emph{AAAI}, 2021.

\end{thebibliography}

% \nextpage
\appendix
\section{Causal Inference: Problem Setup and Challenges}
\label{app:ci}
A dataset 
$
    \mathcal{D}=\left\{ \ [ x_i,\,t_i,\, y_i ]\ \right\}_{i=1}^N
$
used for treatment effect estimation has the following format:
for the $i^{th}$ instance (\eg patient), 
we have some context information $x_i \in \mathcal{X} \subseteq \Real^{K}$ 
(\eg age, BMI, blood work, etc.), 
the administered treatment $t_i$ chosen from a set of treatment options $\mathcal{T}$ 
(\eg \{$0$:\,medication, $1$:\,surgery\}),
and the 
associated % respective 
observed outcome $y_i \in \mathcal{Y}$ 
(\eg survival time: $\mathcal{Y} \subseteq \Real^+$) 
as a result of receiving treatment $t_i$.

Note that $\mathcal{D}$ only contains the outcome of the administered treatment 
(aka~\emph{observed} outcome: $y_i$),
but not the outcome(s) of the alternative treatment(s) 
(aka~\emph{counterfactual} outcome(s);
that is, $y_i^{t}$ for $t \in \mathcal{T} \setminus \{t_i\}$ 
\footnote{
For the binary-treatment case, we denote the alternative treatment as $\neg t_i = 1-t_i$.
}),
which are inherently \textbf{unobservable}~\citep{holland1986statistics}.
In other words, the causal effect 
% \note{Where defined... connected to counterfactual}
$y_i^1 - y_i^0$
is never observed (\ie missing in any training data)
and cannot be used to train predictive models, 
nor can it be used to evaluate a proposed model.
This makes estimating causal effects a more difficult problem than that of generalization in the supervised learning paradigm.

\begin{figure}%[t]
    \begin{center}
        \subfigure[Randomized Controlled Trial]{\label{fig:rct}\includegraphics[width=.45\columnwidth]{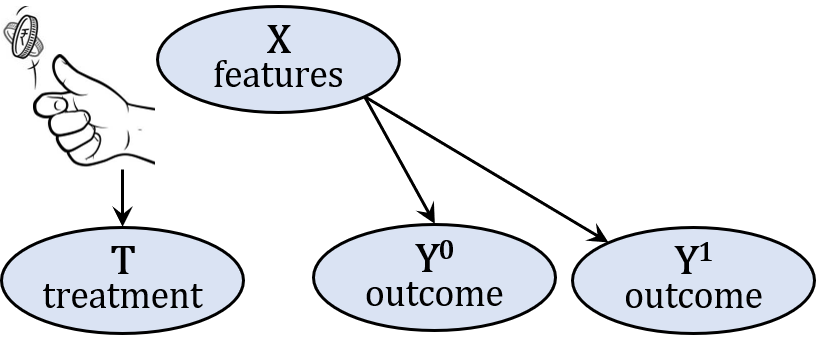}}
        \qquad
        \subfigure[Observational Study]{\label{fig:obs}\includegraphics[width=.45\columnwidth]{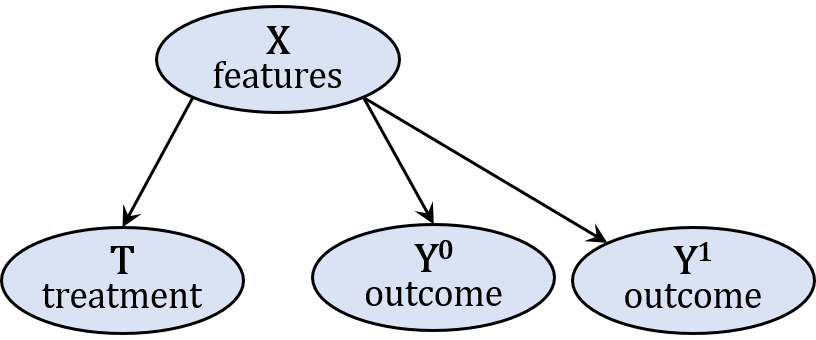}}
    \end{center}
    \caption{
    Belief net structure for randomized controlled trials~(a) and observational studies~(b).
    Here, $Y^0$~($Y^1$) is the outcome of applying $T\!=$~treatment\#0~(\#1) to the individual represented by $X$.
    }
    \label{fig:pgm}
\end{figure}

Pearl~\citep{pearl2009causality} demonstrates that,
in general, causal relationships can only be learned by experimentation (on-line exploration), 
or running a Randomized Controlled Trial (RCT),
where the treatment assignment $T$
does not depend on the individual $X$
-- see Figure~\ref{fig:rct}.
In many cases, however, collecting RCT data is expensive, unethical, or even infeasible.

\begin{figure}[t]
% \begin{wrapfigure}{R}{.47\columnwidth}
    \begin{center}
        % \vskip -.33cm
        \centerline{\includegraphics[width=.65\columnwidth]{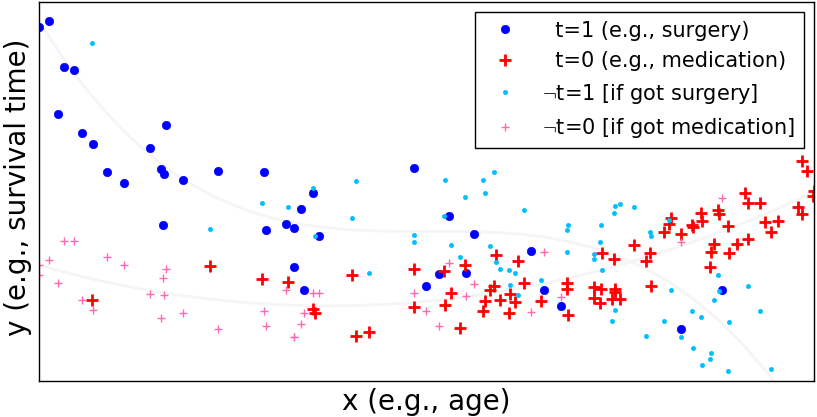}}
        \caption{ 
        An example observational dataset.
        }
        \label{fig:exmpl}
    \end{center}
% \end{wrapfigure}
\end{figure}

A solution is to approximate treatment effects from off-line datasets collected through Observational Studies. 
In such datasets, however, 
the administered treatment $T$ depends on some or all attributes of individual $X$
-- see Figure~\ref{fig:obs}.
Here, as $\CPr{T}{X} \neq \Pr{T}$,
we say these datasets exhibit \textbf{selection bias}~\citep{imbens_rubin_2015}.
Figure~\ref{fig:exmpl} illustrates selection bias in an example (synthetic) observational dataset.
Here, to treat heart disease, a doctor typically prescribes 
    surgery $(t\!=\!1)$ to younger patients~(\textcolor{blue}{\textbf{$\bullet$}})
and 
    medication $(t\!=\!0)$ to older ones~(\textcolor{red}{\textbf{+}}).
Note that instances with larger (resp.,~smaller) $x$ values have a higher chance to be assigned to the $t\!=\!$ 0 (resp.,~1) treatment arm; 
hence we have selection bias.
The counterfactual outcomes 
(only to be used for evaluation purpose) 
are illustrated by 
small fainted 
{\tiny \litecap{\textbf{$\bullet$}}}~({\tiny \pinkcap{\textbf{+}}}) for $\neg t\!=\!$ 1~(0).

\section{Background}
\subsection{M1 and M2 Variational Auto-Encoders}
\label{app:m1m2}
\begin{figure}[t]
    \begin{center}
        \subfigure[M1 model]{\label{fig:m1}\includegraphics[scale=.45]{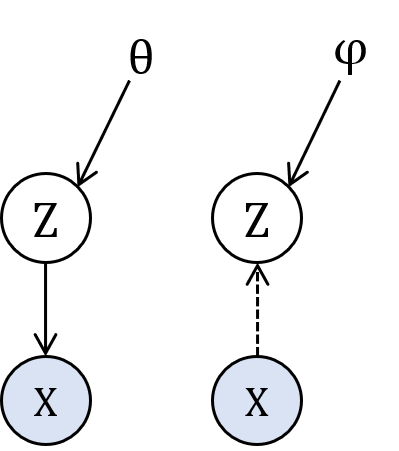}}
        \quad \quad \quad
        \subfigure[M2 model]{\label{fig:m2}\includegraphics[scale=.45]{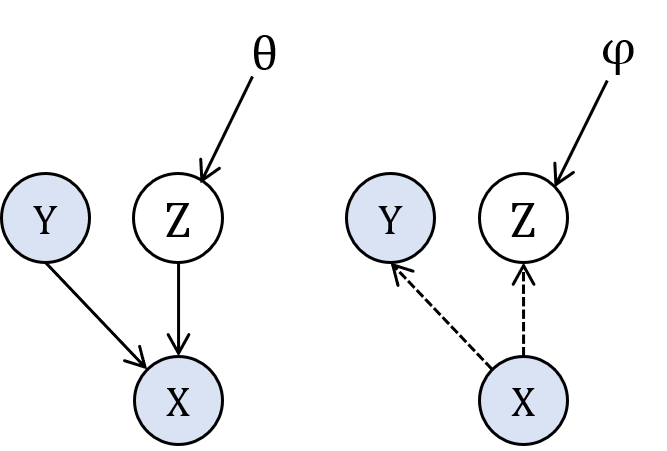}}
        \quad \quad \quad
        \subfigure[M1+M2 model]{\label{fig:m1p2}\includegraphics[scale=.45]{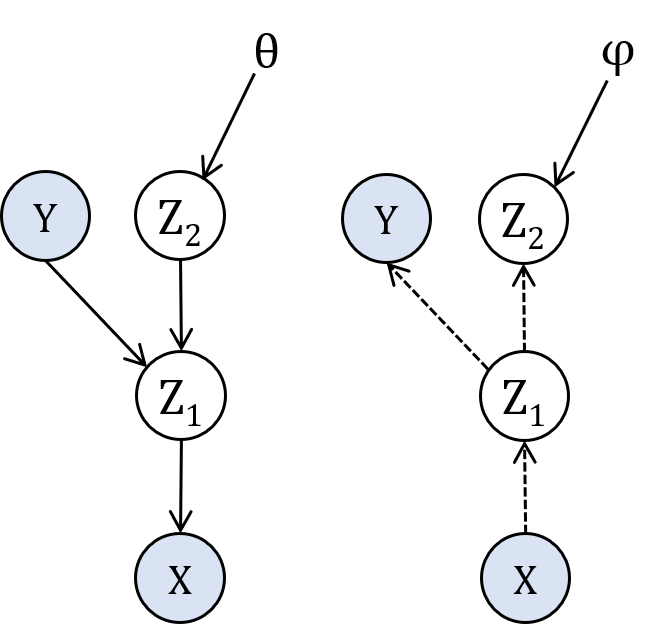}}
    \end{center}
% %\vskip -.15in
    \caption{
        Decoders (parametrized by $\theta$) and encoders (parametrized by $\varphi$) of the M1, M2, and M1+M2 VAEs.
    }
    \label{fig:m1m2}
\end{figure}

As the first proposed model, 
the M1 VAE is the conventional model that is used to learn representations of data \citep{kingma2014auto,rezende2014stochastic}. 
These features are learned from the covariate matrix $X$ only.
Figure~\ref{fig:m1} illustrates the decoder and encoder of the M1 VAE.
Note the graphical model on the left depicts the decoder; 
and the one on the right depicts the encoder,
which has 
% \add[RG]{$X - Z$} 
arrows going the other direction.

Proposed by \citep{kingma2014semi}, 
the M2 model was an attempt to incorporate the information in target $Y$ into the representation learning procedure.
This results in learning representations that separate specifications of individual targets from general properties shared between various targets.
In case of digit generation, 
this translates into separating specifications that distinguish each digit from writing style or lighting condition.
Figure~\ref{fig:m2} illustrates the decoder and encoder of the M2 VAE.

We found that stacking the M1 and M2 models,
as shown in Figure~\ref{fig:m1p2},
produced the best results.
This way, we can first learn a representation $Z_1$ from raw covariates,
then find a second representation $Z_2$, now learning from $Z_1$ instead of the raw data.

%\vskip -.15in

\section{Experimental Setup}
\subsection{Procedure of Generating the Synthetic Datasets}
\label{app:syn}
Given as input 
    the sample size $N$; 
    dimensionalities $[m_\Gamma, m_\Delta, m_\Upsilon] \in ({\cal Z}^{\geq 0})^3$;
    % \note{changed from $Z^{+(3)}$,     as you are including 0.}
    for each factor $L \in \{ \, \Gamma, \Delta, \Upsilon \, \}$,
    the means and covariance matrices $(\mu_L, \Sigma_L)$;
    and
    a scalar $\zeta$ that determines the slope of the logistic curve.
    % \note{Problem with itemize here... ??}
\begin{itemize}%[itemsep=.5pt, topsep=1pt, leftmargin=*]
% \begin{small}
\item For each latent factor $L \in \{ \, \Gamma, \Delta, \Upsilon \, \}$,
    form $L$ by drawing $N$ instances (each of size $m_L$) from $\mathcal{N}(\mu_L, \Sigma_L)$.
    The covariates matrix $X$ is the result of concatenating $\Gamma$, $\Delta$, and $\Upsilon$.
    Refer to the concatenation of $\Gamma$ and $\Delta$ as $\Psi$ 
    and that of $\Delta$ and $\Upsilon$ as $\Phi$ (for later use).
\item For treatment $T$,
    sample $m_\Gamma\!+\!m_\Delta$ tuple of coefficients $\theta$ from $\mathcal{N}(0, 1)^{m_\Gamma\!+\!m_\Delta}$.
    Define the logging policy as $\pi_0(\,t\!=\!1\, |\, z\,)\ =\ \frac{1}{1+\textrm{exp}(-\zeta z)}$, 
    where $z = \Psi \cdot \theta$.
    For each instance $x_i$, sample treatment $t_i$ from the Bernoulli distribution with parameter $\pi_0(\,t\!=\!1 \,|\, z_i\,)$.
\item For outcomes $Y^0$ and $Y^1$,
    sample $m_\Delta\!+\!m_\Upsilon$ tuple of coefficients 
        $\vartheta^0$ 
    and 
        $\vartheta^1$ 
    from $\mathcal{N}(0, 1)^{m_\Delta\!+\!m_\Upsilon}$
    Define 
        $\; y^0 = (\Phi \circ \Phi \circ \Phi + 0.5) \cdot \vartheta^0 / (m_\Delta\!+\!m_\Upsilon) + \varepsilon \;$ 
    and 
        $\; y^1 = (\Phi \circ \Phi) \cdot \vartheta^1 / (m_\Delta\!+\!m_\Upsilon) + \varepsilon$,
    where 
        $\varepsilon$ is a white noise sampled from $\mathcal{N}(0, 0.1)$
    and 
        $\circ$ is the symbol for element-wise product.
% \end{small}
\end{itemize}

\subsection{Hyperparameters}
\label{app:grid}
For all CFR, {DR-CFR}, and {VAE-CI} methods,
we trained the neural networks with 
    3 layers (each consisting of 200 hidden neurons)%
\footnote{
    % \add[R1 \& R2]{
        In addition to this basic configuration,
        we also performed our grid search with an updated number of layers and/or number of neurons in each layer.
        This makes sure that all methods enjoy a similar model complexity. 
    % }
},
    non-linear activation function $\, \mathtt{elu} \,$, 
    regularization coefficient of $\lambda$=1E-4, 
    $\, \mathtt{Adam} \,$ optimizer 
    \citep{kingma2015adam} 
    % [Kingma and Ba, 2015]\note{Why not [xx]?}
    with a learning rate of 1E-3, 
    batch size of 300,
    and 
    maximum number of iterations of $10,000$.
See Table~\ref{tab:hyper} for our hyperparameter search space.
\begingroup
\renewcommand{\arraystretch}{1.25} % Default value: 1
\begin{table}%[H]
\caption{Hyperparameters and ranges}
\label{tab:hyper}
\centering
\begin{tabular}{rl}  
    \toprule
    \textbf{Hyperparameter}             & \textbf{Range}                            \\
    \midrule
    Discrepancy coefficient $\ \alpha$  & \big\{0, 1E\{-3, -2, -1, 0, 1\}\big\}     \\
    KLD coefficient $\ \beta$           & \big\{0, 1E\{-3, -2, -1, 0, 1, 2\}\big\}  \\
    Generative coefficient $\ \gamma$   & \big\{0, 1E\{-5, -4, -3, -2, -1, 0\}\big\}\\
    \bottomrule
\end{tabular}
% %\vskip -.15in
% \note[RG]{Weird to see a set of sets. Perhaps just   1E\{ $-\infty$, -3, .. \},   then explain that  1E$-\infty$ = 0?}
% \note[NH]{I think it would make things complicated.}
\end{table}

\endgroup

\section{Further Results and Discussions}

\subsection{Analysis of the Effect of $\beta=0$}
\label{app:beta0}

Our initial hypothesis in using \hbox{$\beta$-VAE} was that it might help \emph{further} disentangle the underlying factors,
in addition to the other constraint already in place 
({\ie} the architecture as well as the discrepancy penalty).
However, {Figure~\ref{fig:h_b}} suggests that
close-to-zero or even zero $\beta$s also work effectively. 
Our hypothesis is that the \hbox{H-VAE-CI}'s architecture already takes care of decomposing the $\Gamma$, $\Delta$, and $\Upsilon$ factors, without needing the help of a KLD penalty.%
\footnote{
    Therefore, it appears that we can safely drop out the KLD term altogether;
    which can significantly reduce  the model and time complexity.
}

\begin{figure}[h]
% \vskip -.25cm
    \begin{center}
    \includegraphics[scale=.75]{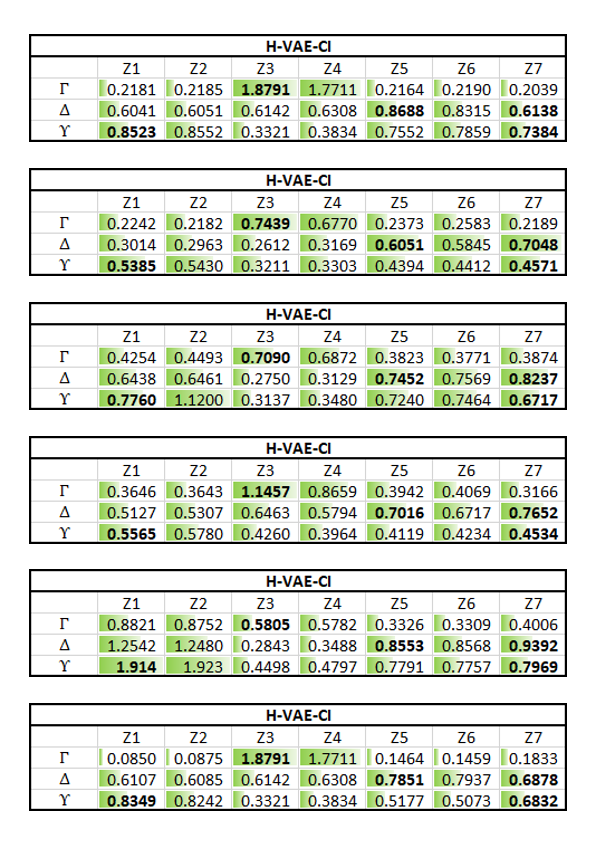}
    \end{center}
% \vskip -.5cm
    \caption{Decomposition tables for \hbox{H-VAE-CI} with $\beta\!=\!0$.}
\label{fig:beta0}
% \vskip -.1cm
\end{figure}

In order to validate this hypothesis, 
we examined the decomposition tables of \hbox{H-VAE-CI} 
(similar to the performance reported in the green table in {Figure~\ref{fig:weight}}) 
for extreme configurations with $\beta=0$ 
and observed that they were all effective at 
decomposing the underlying factors $\Gamma$, $\Delta$, and $\Upsilon$. 
{Figure~\ref{fig:beta0}} shows several of these tables.
This means either of the following is happening:
(i)~\hbox{$\beta$-VAE} is not the best performing disentangling method and other disentangling constraints should be used instead 
    --- {\eg} works of {\citet{chen2018isolating}} and {\citet{lopez2018information}};
    or
(ii)~it is theoretically impossible to achieve disentanglement without some supervision {\citep{locatello2019challenging}}, 
    which might not be possible to provide in this task.
Exploring these options is out of the scope of this paper and is left to future work.

\end{document}